\documentclass[letter, 10 pt, conference]{ieeeconf}  

\IEEEoverridecommandlockouts                              

\usepackage{graphicx}
\usepackage{amsmath}
\usepackage{setspace}
\usepackage{cite}
\usepackage{amssymb}
\usepackage{mathtools}
\usepackage[mathscr]{euscript}
\usepackage{booktabs}
\usepackage{bm}
\usepackage{subcaption}
\usepackage[pdfencoding=auto,psdextra,pagebackref,breaklinks,colorlinks]{hyperref}

\usepackage[capitalize]{cleveref}
\crefname{section}{Sec.}{Secs.}
\Crefname{section}{Section}{Sections}
\Crefname{table}{Table}{Tables}
\crefname{table}{Tab.}{Tabs.}

\newtheorem{theorem}{Theorem}[section]
\newtheorem{corollary}{Corollary}[section]

\DeclareMathOperator*{\argmin}{argmin} 
\newcommand{\vect}[1]{\mathbf{#1}}
\newcommand{\fc}{F}

\usepackage[normalem]{ulem}

\title{\LARGE \bf
TTCDist: Fast Distance Estimation From an Active Monocular Camera\\ Using Time-to-Contact
}

\author{Levi Burner$^{1}$,
Nitin J. Sanket$^{2}$,
Cornelia Ferm\"{u}ller$^{3}$,
Yiannis Aloimonos$^{3}$
\thanks{The support of the NSF under awards DGE-1632976 and OISE 2020624 and the USDA NIFA sustainable agriculture system
program under award number 20206801231805 is gratefully acknowledged. }
\thanks{$^{1}$ Corresponding author. Perception and Robotics Group, Electrical and Computer Engineering, University of Maryland, College Park,
        {\tt\small lburner@umd.edu}}
\thanks{$^{2}$ Robotics Engineering, Worcester Polytechnic Institute
        {\tt\small nsanket@wpi.edu}}
\thanks{$^{3}$ Perception and Robotics Group, University of Maryland Institute for Advanced Computer Studies, University of Maryland, College Park
        {\tt\small \{fer, jyaloimo\}@umiacs.edu }}
}

\begin{document}

\maketitle
\thispagestyle{empty}
\pagestyle{empty}

\begin{abstract}
Distance estimation from vision is fundamental for a myriad of robotic applications such as navigation, manipulation, and planning. Inspired by the mammal's visual system, which gazes at specific objects, we develop two novel constraints relating time-to-contact, acceleration, and distance that we call the \textit{$\tau$-constraint} and \textit{$\Phi$-constraint}. They allow an active (moving) camera to estimate depth efficiently and accurately while using only a small portion of the image. The constraints are applicable to range sensing, sensor fusion, and visual servoing.

We successfully validate the proposed constraints with two experiments. The first applies both constraints in a trajectory estimation task with a monocular camera and an Inertial Measurement Unit (IMU). Our methods achieve 30-70\% less average trajectory error while running 25$\times$ and 6.2$\times$ faster than the popular Visual-Inertial Odometry methods VINS-Mono and ROVIO respectively. The second experiment demonstrates that when the constraints are used for feedback with efference copies the resulting closed loop system's eigenvalues are invariant to scaling of the applied control signal. We believe these results indicate the $\tau$ and $\Phi$ constraint's potential as the basis of robust and efficient algorithms for a multitude of robotic applications.

\end{abstract}

\begin{figure}
  \centering
  \begin{subfigure}{0.9\linewidth}
    \includegraphics[width=1.0\linewidth]{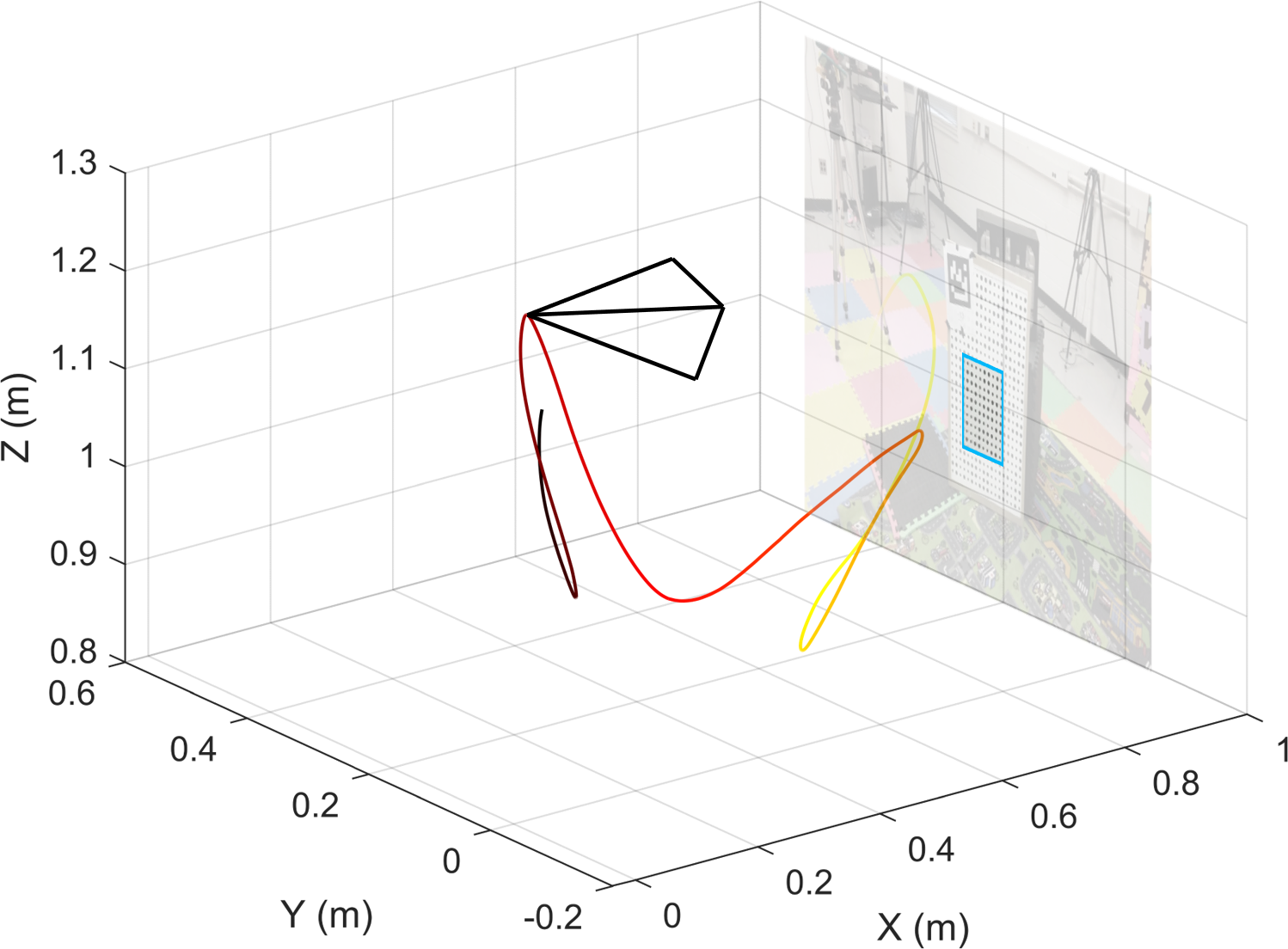}
    \label{fig:scene_1}
  \end{subfigure}
  \hfill
  \begin{subfigure}{0.9\linewidth}
    \includegraphics[width=1.0\linewidth]{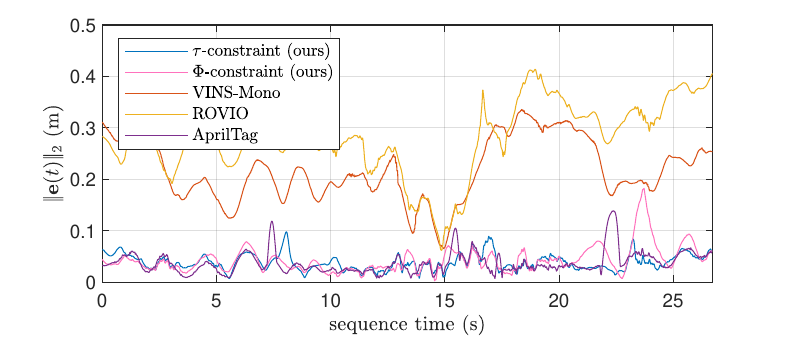}
    \label{fig:scene_5}
  \end{subfigure}
  \hfill
  \caption{Top: Five seconds of Sequence 9's camera trajectory along with the fixated scene patch used to estimate 3D distance using a monocular camera and an IMU. Sequence time is color coded with the hot colormap. Bottom: Instantaneous Euclidean error for Sequence 9 of our methods as well as VINS-Mono, ROVIO, and AprilTag 3. Ground truth is measured by a motion capture system.}
  \label{fig:results_banner}
\end{figure}


\section*{SUPPLEMENTARY MATERIAL}
The experimental data, code, extended proofs, and video are available at \url{prg.cs.umd.edu/TTCDist}.

\section{INTRODUCTION}

Early researchers in computer vision were fascinated by living beings' ability to control their movement in order to gather information about their environment. The process was named ``Active Perception'' \cite{aloimonos1988active, bajcsy1988active,ballard1991animate,fermuller93,fermuller1995vision,nitinthesis}, and numerous mechanisms for utilizing activeness have been developed since. In this work, we focus on the active process of fixation based time-to-contact estimation and using it to measure distance.


Because active vision systems are usually moving in some way, they must accelerate to produce changes to this movement. Despite this, utilizing observer acceleration (measured using an Inertial Measurement Unit or IMU) 
to facilitate visual computations has not received much attention in the Active Vision literature. To fill this gap, we introduce two mathematical constraints relating (a) time-to-contact ($\tau$) (the ratio of camera velocity to scene distance), (b) the relative size of a planar patch over time, with (c) scene depth and (d) observer acceleration. These constraints make it possible for an active monocular camera to estimate scene depth by accelerating in any direction. We call these constraints the $\tau$-constraint and the $\Phi$-constraint.

We demonstrate the utility of the constraints in a series of experiments involving fixation on a single object (tracking) while a monocular camera accelerates.
The method is naturally suited for object-centered representations which 
 have been argued to be useful for a variety of tasks \cite{ballard1991animate, bajcsy2018revisiting,fermuller1992tracking,mishra2009active}. 

Further, we demonstrate a novel mathematical property of the constraint. When efference copies of the control signal (defined in psychology as an internal copy of the movement producing signal) are used in the place of acceleration in the $\tau$ or $\Phi$ constraint, the closed loop dynamics become invariant to scaling of the applied control signal. Informally speaking, this means the weight of the robot, or the strength of the motors, can change without influencing stability.

The key concept of fusing inertial measurements with camera observations has been extensively studied in the fields of Structure from Motion (SfM) and Simultaneous Localization and Mapping (SLAM). Traditionally, this fusion has been achieved in Visual Inertial Odometry (VIO) frameworks using a series of Bayesian filters or Factor graphs \cite{indelman2013information,santoso2016visual}.

It is important to stress that our implementation is not a fully-functional VIO implementation, and is only intended to empirically demonstrate the efficacies of the $\tau$ and $\Phi$ constraints for absolute position estimation and control. Nevertheless, our constraints naturally result in a trajectory estimate. For this reason, we compare our method with the popular state-of-the-art VIO methods such as VINS-Mono\cite{qin2018vins} and ROVIO \cite{bloesch2015robust} as well as the fiducial marker based pose estimation method, AprilTag 3~\cite{krogius2019iros}. We also compare our results with millimeter accurate ground truth from a Vicon motion capture system.




A list of our contributions follows:

\begin{itemize}
    \item A closed form solution for the 3D position of the camera given time-to-contact or depth-to-scale and acceleration.
    \item A computationally efficient position estimation method utilizing the $\tau$ or $\Phi$ constraint based on the closed form.
    \item Comparisons against the popular VIO methods, VINS-Mono and ROVIO, as well as AprilTag 3, to show the efficacy of the novel constraint in real-world settings. 
    \item A corollary (resulting from the closed form) and experiment showing that our constraints, when used with efference copies in a closed loop, make the system invariant to scaling of the control signal.
\end{itemize}



\section{Related Work}
A multitude of prior works from both the computer vision and robotics literature deal with time-to-contact and the fusion of camera and IMU measurements to obtain relative camera pose (odometry). However, to the best of our knowledge, none of these works provided closed form solutions for the estimation of distance from time-to-contact and inertial measurements nor for distance from the apparent size of a tracked planar patch and inertial measurements.

\subsubsection*{Time-to-contact}
One of the earliest works to discuss how time-to-contact $\tau$ can be used in control tasks came from psychology. \cite{lee1976theory} provided an analysis of how time-to-contact, $\tau$, obtained from vision, could be used by drivers to control braking a vehicle. The idea was generalized to other perception modalities into a ``General Tau Theory'' in \cite{leegeneral2009}.

Because of its intuitive formulation, $\tau$ has also been the subject of many studies in robotics. \cite{Sikorski2021} showed how to land a spacecraft using event cameras by computing $\tau$ from the divergence of optical flow. \cite{walters2021evreflex} fuses information from a depth camera and $\tau$ to compute ``time-to-impact'' which can  in-turn be used to dodge dynamic obstacles.
In robotics, most methods that perform optical flow based control use initial height estimates either implicitly or explicitly, as remarked in \cite{ho2017distance}. To this end, \cite{ho2017distance} proposed to fuse control effort and time-to-contact with an extended Kalman filter which estimated depth. Another strategy exploits the instability that manifests at certain heights when performing direct $\tau$ control with fixed gain feedback to estimate depth \cite{de2015distance}. In the context of self-driving cars, BinaryTTC \cite{badki2021binary} proposed a network to predict per-pixel $\tau$. Recently, EV-Catcher caught fast-moving objects using a network to predict time-to-contact and the current position of a moving object \cite{Wang2022EVcatch}.
Finally, it is important to note that $\tau$ can be efficiently computed when the scene under consideration is planar \cite{horn2007time}.

\begin{figure*}[t]
  \centering
  \includegraphics[width=1.0\linewidth]{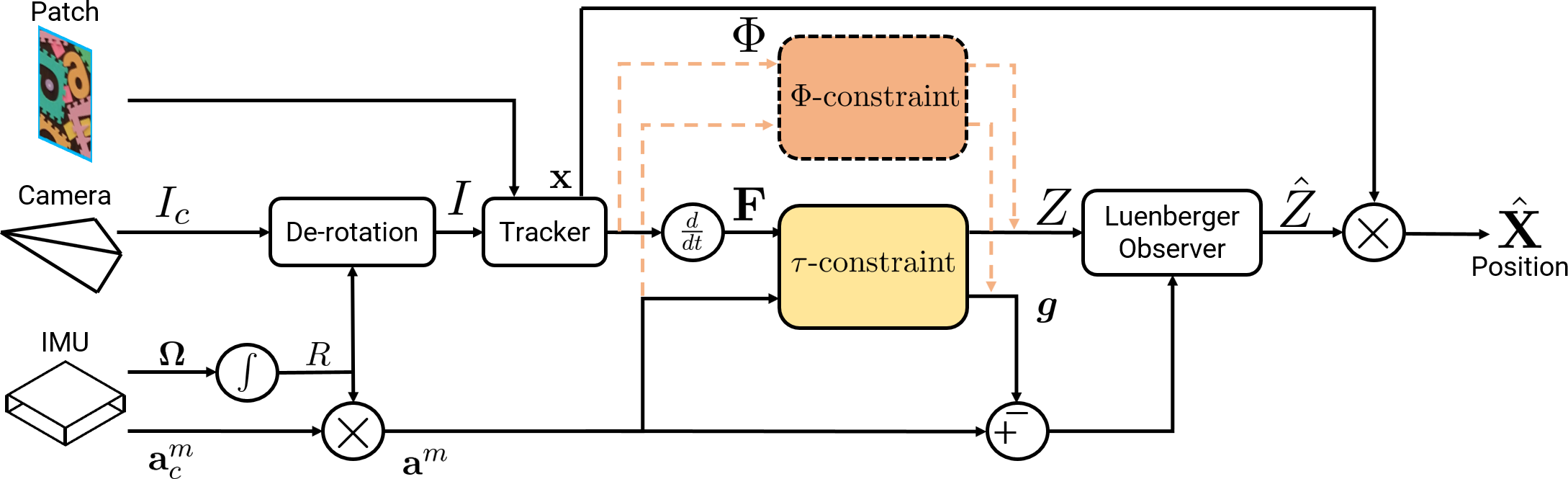}
   \caption{System overview of our method to estimate camera position using the $\tau$-constraint. The $\Phi$-constraint uses an identical process except that the derivative of $\Phi$ is not taken. Only one constraint is used at a time as indicated by the dashed arrows and outlines.}
   \label{fig:systemdiagram}
\end{figure*}

\subsubsection*{Visual Inertial Odometry (VIO)}
One of the earliest works for real-time VIO fused sparse feature tracks with IMU measurements using a Multi-State Constraint Kalman Filter \cite{mourikis2007multi}. This pipeline was made more robust by ROVIO \cite{bloesch2015robust} which fused photometric consistency of patches with IMU measurements in an Iterative Extended Kalman Filter. ROVIO relies on tracking several planar (affine warpable) patches distributed across the field of view and uses approximations to implement the iterated EKF. To this end, VINS-Mono \cite{qin2018vins} introduced a point-based nonlinear sliding window estimator with an initialization-free formulation where the points are assumed to be distributed across the field of view. These methods are supported by a proof that tracked points and inertial measurements allow pose to be recovered in closed form \cite{martinelli2011vision, martinelli2014closed}. In contrast, our constraints admit closed form solutions for distance to a single patch of unknown size.


Recent deep learning based VIO methods have achieved better accuracy than classical approaches. VINet \cite{clark2017vinet} presented the first supervised method to estimate VIO using a CNN + LSTM architecture. This was later improved by DeepVIO \cite{han2019deepvio} using supervision from a stereo camera. The limiting factor of these approaches is a lack of speed and generalization across various compute platforms \cite{sanket2021prgflow}. As will be shown, the $\tau$-constraint can be formulated as a loss, which may be interesting for future work in deep VIO. We will present a derivation of the proposed constraints next.

\section{Derivation of the $\tau$ and $\Phi$ constraint}



We develop a system to use the $\Phi$ and $\tau$ constraints with a calibrated monocular camera and an aligned 6-DoF IMU that measures acceleration and angular velocity. Referring to \cref{fig:systemdiagram}, we compose the incoming frames with a rotational warp function from the IMU. Then we ``fixate on'' (track) a planar patch in the rotation-compensated images using an affine homography. A history of the parameters of the affine homography are then used to estimate distance with our proposed $\Phi$ and $\tau$ constraints. Finally, we filter the predictions using a Luenberger observer (explained in \cref{sec:fusion}) to obtain the final trajectory estimates.

In the following derivation, we first define frequency of contact, $\vect{F}$, and explain its relation to $\Phi$. Then we show how $\vect{F}$ and $\Phi$ can be measured from an affine homography. Finally, we related $\vect{F}$ and $\Phi$ to acceleration and depth which results in the $\tau$ and $\Phi$ constraints respectively.

\subsection{Definition of $\tau$, frequency-of-contact $\mathbf{F}$, and $\Phi$}

Time-to-contact or $\tau$ is defined as $Z/\dot{Z}$ (ratio of depth/distance to velocity). Below, we generalize $\tau$ to all three dimensions and define frequency of contact as

\begin{equation}\label{eq:frecontact}
\vect{\fc} \coloneqq \dfrac{\dot{\vect{X}}}{Z}.
\end{equation}

\begin{figure}[b]
  \centering
  \includegraphics[width=0.8\linewidth]{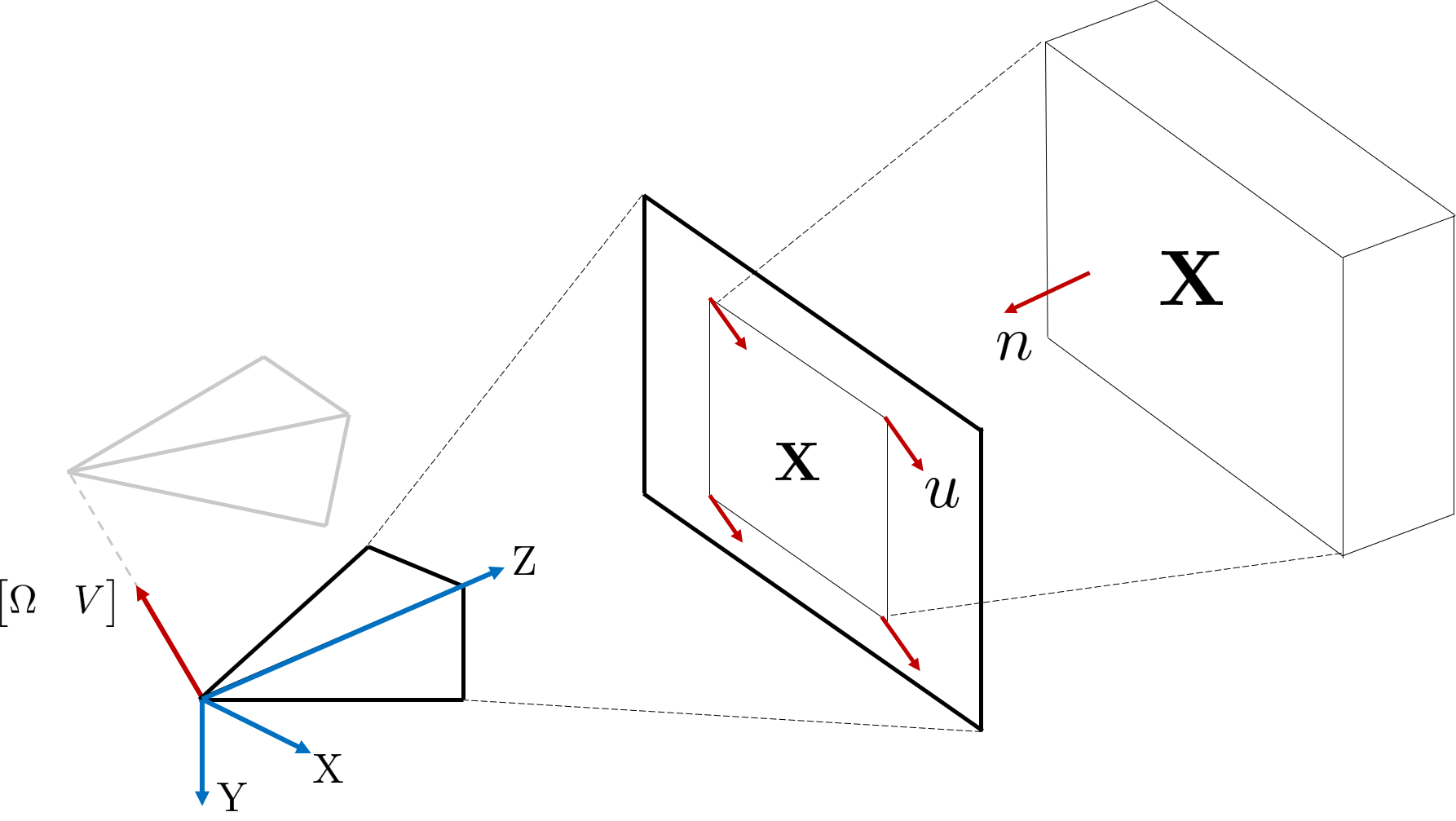}
   \caption{By the pinhole model, a scene point $\vect{X}$ is projected to image point $\vect{x}$. The camera ``perceives'' movement of $\vect{x}$ as optical flow $\vect{u}$.}
   \label{fig:scenemotion}
\end{figure}

$\vect{X} = [X, Y , Z]^T$ is the position of a point as in \cref{fig:scenemotion}.

Frequency of contact is the number of times per second a constant speed point will reach an axis.
Unlike time-to-contact, frequency-of-contact is only ill-defined when $Z=0$, which just means the object cannot be seen. The third component of $\vect{F}$, $F_Z$, is equal to $1/\tau$.

Frequency-of-contact is directly related to $\Phi$ (depth and translation to scale). To show this, consider that $\vect{F}$ defines the linear time varying system $\dot{\vect{X}} = \vect{F} Z$ which has the solution

\begin{equation}
\label{eqn:phidefn}
\vect{X}(t) = \underbrace{\begin{bmatrix}
1 & 0 & \int_{0}^t \fc_X(\lambda) \Phi_{\fc_Z}(\lambda) d\lambda \\
0 & 1 & \int_{0}^t \fc_Y(\lambda) \Phi_{\fc_Z}(\lambda) d\lambda \\
0 & 0 & \Phi_{\fc_Z}(t)
\end{bmatrix}}_{\Phi(t) \coloneqq} \vect{X}_0,
\end{equation}

where $\Phi_{F_Z}$ is the solution to the ODE $\dot{Z} = F_Z Z$ when $Z_0 = 1$, i.e.\
$
\Phi_{F_Z}(t) = \exp\left(\int_{0}^t \fc_Z(\lambda) d\lambda\right)
$ \cite{Hespanha2018}.


Next, we use a tracked planar patch to estimate $\vect{F}$ and $\Phi$.

\subsection{Estimating $\tau$ and $\Phi$ from an Affine Homography}
As illustrated in \cref{fig:scenemotion}, a point in a scene $\vect{X}$ is projected to a point on the image $\vect{x}$ according to the pinhole model $\vect{x} = \vect{X} / Z$. Without loss of generality, we assume that the camera intrinsic matrix $K$ is identity. If we are tracking the points on a plane that is parallel to the imaging plane and the camera translates without rotating, then an affine homography relates all points on the planar patch in the current frame to their positions in the first frame and is given by


\begin{equation}
\vect{x}(t) =
\underbrace{\begin{bmatrix}
    Z_0/Z & 0 & (X - X_0) / Z\\ 
    0 & Z_0/Z & (Y - Y_0) / Z\\
    0 & 0 & 1
\end{bmatrix}}_{A \coloneqq} \vect{x}(0).
\end{equation}

The definition of $\Phi$ in \cref{eqn:phidefn} allows us to write $\vect{X}-\vect{X}_0 = (\Phi - I)\vect{X}_0$ which reveals that the components of $\Phi - I$ are determined by the elements of $A$. In our implementation, we estimate the affine homography $A$ using the inverse Lucas-Kanade method. As studied in \cite{baker2004lucas}, the patch must have sufficient texture to ensure convergence. We also compose the affine homography with the rotation from inertial measurements (as shown in \cref{fig:systemdiagram}).

To obtain $\vect{F}$ on the tracked patch from the affine homography we consider the optical flow $\vect{u}_{\vect{x}}$ as a function of $\vect{x}$


\begin{equation}
\label{eqn:bdefn}
    \vect{u}_{\vect{x}} = \frac{d \vect{x}(t)}{dt} =
\dot{A} A^{-1} \vect{x}
=
-
\underbrace{
\begin{bmatrix}
    \dot{Z}/Z & 0 & \dot{X}/Z\\ 
    0 & \dot{Z}/Z & \dot{Y}/Z\\
    0 & 0 & 0
\end{bmatrix}}_{B \coloneqq }
\vect{x}.
\end{equation}

Thus, $B$'s terms are equal to frequency-of-contact $\vect{F}$. In practice, we fit an affine homography to a slightly slanted plane. In that case, it is more appropriate to consider the affine flow parameters in $B= [b_{i,j}]$ as corresponding to the linear terms of optical flow due to a planar surface. Then, if  $\eta \coloneqq ((b_{2,1} b_{1,3} / b_{2,3}) - b_{1,1})$, the frequency-of-contact is

\begin{equation}
\label{eqn:affinefreq}
\dfrac{\dot{\vect{X}}}{Z} = \begin{bmatrix}
b_{2,1} b_{1,3}/b_{2,3} &  b_{1,2} & b_{1,3} \\
b_{2,1} & b_{1,2} b_{2,3}/b_{1,3} & b_{2,3} \\
\eta(b_{2,1}/b_{2,3}) & \eta(b_{1,2}/b_{1,3}) & \eta
\end{bmatrix}
\vect{x}.
\end{equation}

Next, our constraints are derived from the above relations.

\subsection{The $\tau$-constraint and the $\Phi$-constraint}

Now, we will relate $\vect{F}$ and $\Phi$ to depth and acceleration which results in the $\tau$ and $\Phi$ constraints respectively. The constraints relate the initial conditions of the linear time varying system defined by time-to-contact and the linear time invariant system defined by acceleration.

By the fundamental theorem of calculus, $\vect{X}$ is given by
\begin{equation}
\label{eqn:ltisolution}
\vect{X}(t) - \vect{X}_0 = t \dot{\vect{X}}_0 + \underbrace{\int_{0}^t \left( \int_{0}^{\lambda} \ddot{\vect{X}} (\lambda_2) d\lambda_2 \right) d\lambda}_{\mathscr{J}\{\ddot{\vect{X}}\}(t) \coloneqq }.
\end{equation}

Where $\mathscr{J}\{\vect{f}\}(t) : L_p^n \times \mathbb{R} \to \mathbb{R}^n$ is an operator returning the double integral of a vector of functions \cite{birmanlec}.

Since $\vect{X}-\vect{X}_0 = (\Phi - I)\vect{X}_0$ we can make a substitution which results in the $\Phi$ constraint.\\[-5pt]

\begin{theorem}[$\Phi$-constraint]
If $\Phi$ and $\ddot{\vect{X}}$ are known $\forall t$ in a closed time interval and $Z \geq \epsilon > 0, \forall t$, then the following linear constraint between initial depth $Z_0$, $\Phi$, and acceleration $\ddot{\vect{X}}$ holds for each point on the planar scene patch
\begin{equation}\tag{$\Phi$-constraint}
\left(\Phi(t) - I \right) \begin{bmatrix}0 \\ 0 \\ Z_0\end{bmatrix} - t \dot{\vect{X}}_0 = \mathscr{J}\{\ddot{\vect{X}}\}(t).
\end{equation}
\end{theorem}

Using the $\Phi$-constraint for position estimation requires determining four unknowns, $Z_0$ and $\dot{\vect{X}}_0$. Using the fact that $\left(\Phi(t) - I \right) \propto t$ if and only if $\ddot{\vect{X}} = 0, \forall t$ makes it possible to show that when $\Phi(t)$ and $\mathscr{J}\{\ddot{\vect{X}}\}(t)$ are known over a time interval, then the linear system defined by the $\Phi$-constraint determines $Z_0$ and $\dot{X}_0$ if and only if $\exists \ t$ between the two times $ \text{s.t.} \  \ddot{\vect{X}} \neq 0$. A proof is in the supplementary material.

Since $\dot{\vect{X}}_0 = \vect{F}(0) Z_0$, the $\tau$-constraint follows directly.\\[-5pt]

\begin{theorem}[$\tau$-constraint]
If $\vect{F}$ and $\ddot{\vect{X}}$ are known $\forall t$ in a closed interval and $Z \geq \epsilon > 0, \forall t$, then the following linear constraint between depth $Z_0$, frequency-of-contact $\vect{F}$, and acceleration $\ddot{\vect{X}}$ holds for each point on the planar scene patch
\begin{equation}\tag{$\tau$-constraint}
\underbrace{\left(\Phi(t) - I - t\begin{bmatrix} 0 & 0 & \vect{F}(0)\end{bmatrix}\right)}_{E(t) \coloneqq}\begin{bmatrix}0 \\ 0 \\ Z_0\end{bmatrix} = \mathscr{J}\{\ddot{\vect{X}}\}(t).
\end{equation}
\end{theorem}



$E(t)$ is the ratio between positional change due to acceleration and the depth $Z_0$. Intuitively, it is the \textit{``action's effect''}.

It is proven in the supplementary material that determining $Z_0$ is well posed if and only if $\ddot{\vect{X}}$ is non-zero at some time.

The $\Phi$-constraint and $\tau$-constraint are closely related. The $\Phi$-constraint considers the position and area of an object in the image, and the $\tau$-constraint considers its velocity and rate of change of size. Since $\vect{F}$ determines $\Phi$ by integration, it is reasonable to use either the $\tau$ or $\Phi$ constraint when $\vect{F}$ is available. However, using the $\tau$-constraint when only $\Phi$ is available is difficult because $\Phi$ must be numerically differentiated to get $\vect{F}$, which can introduce significant noise.

Next, we discuss estimating depth using our constraints.

\section{Fusing Inertial Measurements with $\tau$ and $\Phi$}
\label{sec:fusion}

To tightly couple IMU measurements and frequency-of-contact from a camera in a sliding window manner, we set up a least squares problem over the constraints. The solution is the depth of a point in the scene and the gravity direction.

As illustrated in \cref{fig:systemdiagram},
we estimate a rotation matrix, $R$, using the gyroscope.
Using this matrix allows all measurements to be rotated into the orientation of the initial camera frame, and thus over a short time period, the method can be considered rotation invariant. Thus, we use $R$ to rotate the measured acceleration $\vect{a}^m_c(t)$ back to a fixed frame, $\vect{a}^m(t) = R(t) \vect{a}^m_c(t)$. The IMU measures the resultant of gravitational acceleration and linear acceleration. Thus, $\vect{a}^m(t) = -\ddot{X} + \bm{g}$.

\subsection{Efficient Computation of Depth ($Z$ Distance)}
Let us suppose a history of $\vect{a}^m(t)$ and $\vect{\fc}(t)$ measurements are available over a time interval $[0, T]$. 
Now, without loss of generality, we consider the problem along only the $Z$ axis. The problem for the $\tau$-constraint can be written as 

\begin{equation}\label{eqn:zproblem}
    \argmin_{Z_0, g_Z} \int_0^T \left({E_Z Z_0 + \mathscr{J}\{a^m_Z + g_Z}\}\right)^2 dt.
\end{equation}

The problem for the $\Phi$-constraint is given by

\begin{equation}\label{eqn:zproblemphi}
    \argmin_{Z_0, \dot{Z}_0, g_Z} \int_0^T \left( {\left(\Phi_{F_Z} - 1\right) Z_0 - r \dot{Z}_0 +  \mathscr{J}\{a^m_Z + g_Z}\} \right)^2 dt,
\end{equation}

where $r(t) \coloneqq t$ is the ramp function.

In both cases, the problem is the same for the $X$ or $Y$ axis up to a change of subscripts and the initial velocity.

It is proven in the supplementary material that \cref{eqn:zproblem} and \cref{eqn:zproblemphi} are well-posed if and only if the measured acceleration, $a^m_z$, is not constant for the entire time interval.

Thus, given motion along an axis, depth and gravitational acceleration can be recovered efficiently by solving a linear least squares problem based on either constraint.
The estimates have noise and so we discuss filtering them next.

\subsection{Filtering of Depth}
Now, we extend the above derivation for trajectory estimation. Since \cref{eqn:zproblem} estimates $g_z$ and $Z$, we can set up a Luenberger observer \cite{Hespanha2018} to filter the trajectory estimate. Let $\hat{Z}$ and $\hat{\dot{Z}}$ be the estimated quantities, then
\begin{equation}
    \begin{bmatrix}
        \dot{\hat{Z}} \\
        \dot{\hat{\dot{Z}}} \\
    \end{bmatrix}
    =
    \begin{bmatrix}
        \hat{\dot{Z}} \\
        a^m_z - g_z \\
    \end{bmatrix}
    + L \begin{bmatrix}
            Z - \hat{Z} \\
            \dot{Z} - \hat{\dot{Z}}\\
        \end{bmatrix}.
\end{equation}


When multiple estimates for $Z$ are available from motion along multiple axes, the results are averaged. When $\hat{Z}$ and $g_z$ are not available, due to lack of acceleration, then dead reckoning is used by applying $\vect{F}$ or $\Phi$ to the latest $\hat{Z}$.

Finally, we recover the three dimensional trajectory by multiplying $\hat{Z}$ with the current image location of the center of the planar patch: $    \begin{bmatrix}
    \hat{X} &
    \hat{Y} &
    \hat{Z}
    \end{bmatrix}^T
    \coloneqq
    \vect{x}
    \hat{Z}
$. Next, we explain how to substitute control effort for acceleration.

\section{Closed Loop control using Efference Copies}

Since the control effort in robotics often corresponds to acceleration up to some scale, it is interesting to consider what happens when control effort, $\vect{u}$, is substituted for $\ddot{\vect{X}}$ when solving \cref{eqn:zproblem}. Control effort and optical flow are often referred to as $\vect{u}$. So we use $\vect{u}_{\vect{x}}$ for the flow at pixel $\vect{x}$.

This substitution leads to an interesting property.\\[-5pt]

\begin{corollary}
\label{thm:invariance}
If $\ddot{\vect{X}} = b \vect{u}, b \neq 0$, where $\vect{u}$ is a linear control determined by $\vect{u} = K \hat{\vect{X}}$, then if $\vect{u}$ is substituted for $\vect{a}^m$ in \cref{eqn:zproblem}, the dynamics become invariant to $b$.
\end{corollary}

\begin{proof}
By definition $\ddot{\vect{X}} / b = \vect{u}$, and  the solution to \cref{eqn:zproblem} is linear in $\vect{a}^m$. Thus, using $\vect{u}$ instead of $\vect{a}^m$ results in a state estimate scaled by the inverse of $b$, i.e.\ $\hat{\vect{X}} = \vect{X} / b$.

Further, since $\vect{u} \coloneqq K \hat{\vect{X}}$, we can see that $\ddot{\vect{X}} = b K (\vect{X} / b) = K \vect{X}$. Thus, the dynamics are invariant to $b$.\\[-5pt]
\end{proof}

Practically speaking, this allows changing the strength of a motor, or the weight of a robot, without drastically changing the stability properties of the system. Typically such a change would require re-tuning the control gains $K$.


\section{Experiments}
We designed the experiments to evaluate the $\tau$ and $\Phi$ constraints as well as the invariance property to determine if the constraints are a promising method for future research and applications.

\subsection{Metric Trajectory Estimation}

To test the constraints' ability for trajectory estimation, we created ten sequences with five distinct scenes in an indoor setting. Each scene contains a planar object to fixate on as shown in \cref{fig:results_banner}.
For each scene, two recordings were made, one with an AprilTag and one without. The trajectories feature acceleration often over 2 $\text{m}/\text{s}^2$.

The sequences were recorded with an Intel$^\text{\textregistered}$ RealSense$^\text{TM}$ D435i camera using the built-in IMU and the left grayscale camera \cite{realsense2017}. We used the grayscale camera because the D435i's IMU is hardware time-stamped to the grayscale imager. The D435i captures images at 90 frames per second at 848 $\times$ 480 px. resolution. The IMU records gyroscope measurements at 400 Hz and the acceleration at 250 Hz.

We tracked the fixated planar patch using gyroscope measurements for rotation stabilization and by continually fitting an affine homography using the ubiquitous patch tracking method from \cite{baker2004lucas}. The frequency-of-contact $\vect{F}$ was then recovered using \cref{eqn:affinefreq}. The affine tracker was initialized by tracking a 100$\times$100 pixel patch sub-sampled to 4000 pixels. While the patch size changes dramatically during fixation, only 4000 pixels are  drawn from each frame.

Then, the $\tau$-constraint was fused with the IMU measurements using a 2 second signal history and a 100 Hz sampling rate using linear interpolation. If the average power of the bias corrected acceleration along an axis was below 2 $\text{m}/\text{s}^2$  the resulting depth from the $\tau$-constraint  was not used. If no observations of depth were available the depth estimate was forward propagated using dead reckoning with $\vect{F}$ or $\Phi$. The Luenberger's gain was set to $L = \text{diag}(2, 20)$ for all the sequences. \textit{Also, it is important to note, that these parameters were not tuned using the sequences considered in this work.}

Our implementation is written in Python 3.8 using standard scientific Python libraries. Numba is used to accelerate critical sections \cite{numba2015}. One thread on an Intel$^\text{\textregistered}$ Core$^\text{TM}$ i7-6820HQ laptop processor was used to perform computations.

The open-source VINS-Mono and ROVIO implementations were used for comparison \cite{qin2018vins, bloesch2015robust, bloesch2017iterated}. VINS-Mono was configured to output poses at 90 Hz (the camera frame rate), however, it produced poses only at 80 Hz. ROVIO was configured to produce poses at 90 Hz.

As another source of comparison, we used the AprilTag 3 \cite{krogius2019iros} library to detect 36h11 tags. The corners were used to solve the Perspective-n-Point problem to recover a tag's location in the current frame. The gyroscope measurements were used to rotate each AprilTag pose measurement back to the fixed orientation used by the constraints.

The ground truth trajectory was measured at 200 Hz using a Vicon motion capture system with 8 Vantage V8 cameras. We align all trajectories to Vicon ground truth and compute the Average Trajectory Error (ATE) as described in \cite{Zhang18iros}.

\begin{equation}
\text{ATE}(\vect{X}, \vect{X}^v) = \left(\frac{1}{N} \sum_{n=0}^{N-1} \Vert \vect{X}_n - \vect{X}^v_n \Vert_2^2 \right)^{1/2}.
\end{equation}


\begin{table}
  \centering
  \caption{Sequence duration (seconds), path length (meters), and each method's accuracy in centimeters of Average Trajectory Error (ATE). Since AprilTag 3 is always the best when it is available we also bold the second best result in such sequences.}
  \resizebox{1.0\columnwidth}{!}{
  \begin{tabular}{@{}lccccc@{}}
    \toprule
  \texttt{Seq.} &  \texttt{1} & \texttt{2} & \texttt{3} & \texttt{4} & \texttt{5} \\
    \midrule
Duration (s) & 15.06 & 26.15 & 32.28 & 36.23 & 16.41\\
Length (m) & 15.73 & 29.65 & 22.21 & 34.75 & 15.63\\
    \midrule
    Method & \multicolumn{5}{c}{ATE (cm)} \\
    \midrule
AprilTag 3 & \textbf{2.80} & - & \textbf{2.67} & - & \textbf{3.76} \\
VINS-Mono & 5.41 & 8.80 & 14.21 & 15.37 & - \\
ROVIO & 7.77 & 9.89 & 11.88 & 33.23 & 29.96 \\
\midrule
$\Phi$-constraint (ours) & \textbf{3.77} & \textbf{5.79} & \textbf{7.60} & \textbf{7.32} & \textbf{7.40} \\
$\tau$-constraint (ours) & 8.07 & 6.91 & 12.33 & 10.21 & 16.82 \\
    \midrule
    \midrule
  \texttt{Seq.} &  \texttt{6} & \texttt{7} & \texttt{8} & \texttt{9} & \texttt{10}\\
    \midrule
Duration (s) & 16.27 & 8.02 & 32.15 & 26.73 & 40.08\\
Length (m) & 15.78 & 7.30 & 26.75 & 21.39 & 35.37\\
    \midrule
    Method &  \multicolumn{5}{c}{ATE (cm)} \\
    \midrule
AprilTag 3 & - & \textbf{0.65} & - & \textbf{2.48} & - \\
VINS-Mono & 6.10 & 1.15 & 18.45 & 13.07 & 4.34 \\
ROVIO & \textbf{2.84} & \textbf{0.69} & 3.93 & 16.62 & 3.79 \\
\midrule
$\Phi$-constraint (ours) & 5.71 & 1.42 & \textbf{3.28} & 2.86 & \textbf{2.42} \\
$\tau$-constraint (ours) & 7.21 & 10.70 & 4.32 & \textbf{2.38} & 3.24 \\
    \bottomrule
  \end{tabular}}
  \label{tab:atecompare}
\end{table}

Here $\vect{X}^v_n$ is themotion capture systems n'th position estimate and $\vect{X}_n$ is the n'th estimated position.


\subsection{Closed Loop Stability Invariance Property}

\begin{figure}
  \centering
  \begin{subfigure}{0.63\linewidth}
    \includegraphics[width=1.0\linewidth]{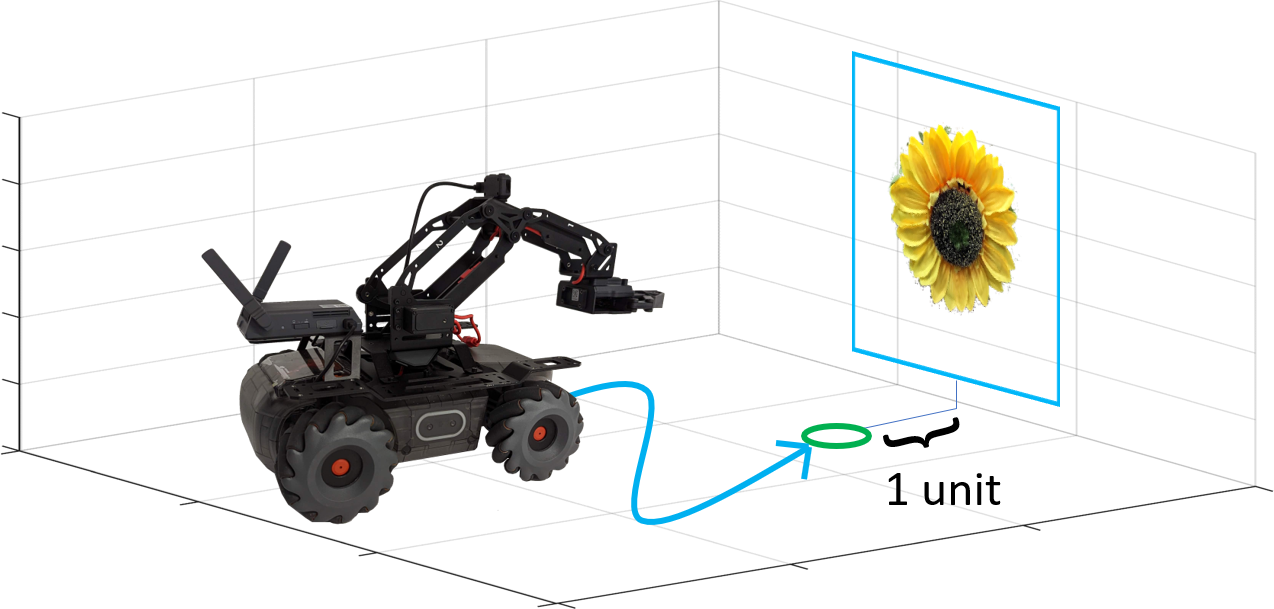}
    \label{fig:djiexperiment}
  \end{subfigure}
  \begin{subfigure}{0.207\linewidth}
    \includegraphics[width=1.0\linewidth]{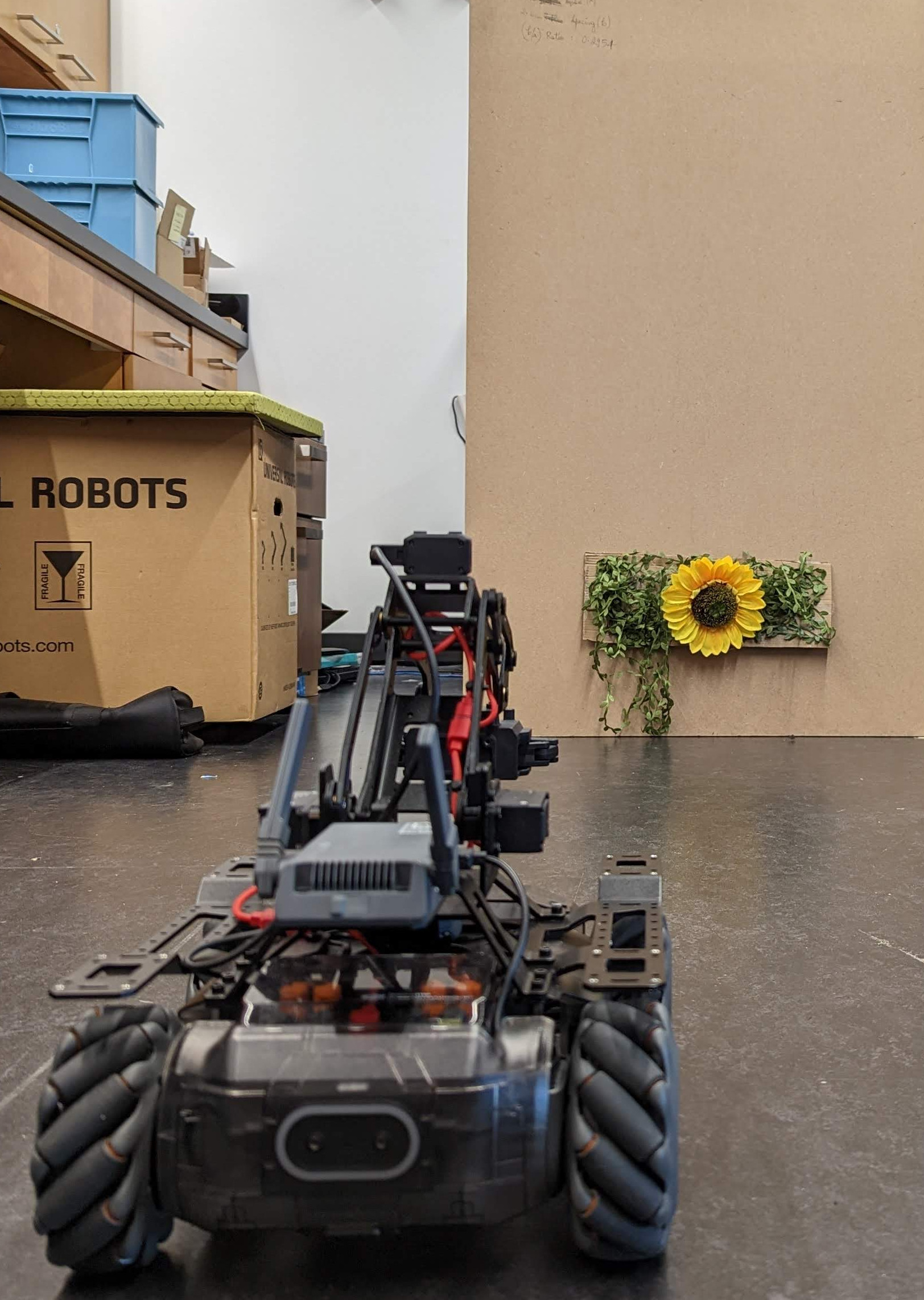}
    \label{fig:djiexperimentcropped}
  \end{subfigure}
  \hfill
  \begin{subfigure}{0.9\linewidth}
    \includegraphics[width=1.0\linewidth]{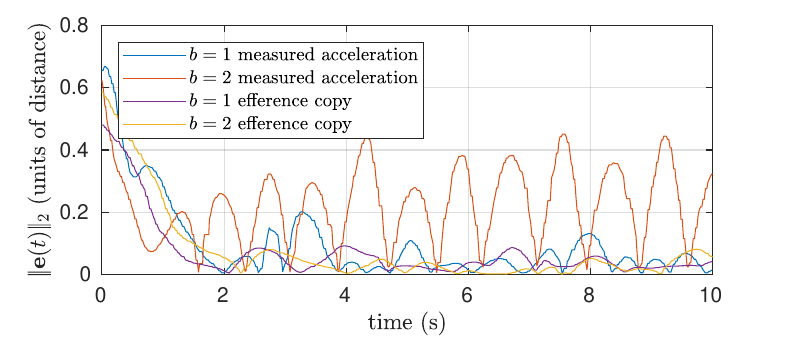}
    \label{fig:djierror}
  \end{subfigure}
  \hfill
   \caption{Top left: The invariance property is tested by using the constraint to navigate to a unit distance from a visual target. Top right: The experimental setup. Bottom: The normed error between the robot's position and the target position. When efference copies are used, the system is stable despite changes to the unknown gain $b$. When using measured acceleration, the system destabilizes when $b$ is set to 2. Distance is in meters for trials using measured acceleration and units of effort when using efference copies.}
   \label{fig:djiexperimentsfull}
\end{figure}

To test the invariance property, we implement a closed loop controller on a DJI$^\text{\textregistered}$ RoboMaster$^\text{TM}$ robot pictured in Fig. \ref{fig:djiexperimentsfull}. The robot's goal is to reach a fixed distance from a pre-determined visual target, where distance is in meters for trials using measured acceleration and units of effort when using efference copies. By adjusting a static gain $b$ applied to the control signal we can test if using efference copies in the $\Phi$-constraint prevents the control system from exhibiting unstable behavior when $b$ is increased dramatically. Four trials were run. In two of these acceleration from the onboard IMU was fed to the $\Phi$-constraint whose estimates were then used for closed loop control. In the other two trials, efference copies were fed to the $\Phi$-constraint, meaning that $\vect{u}$ was used instead of the measured acceleration. For each group of two experiments, one was run with the actuator gain set to 1 (which was the value the $K$ matrix was tuned for). In the second run, the actuator gain was set to 2, which doubled all control signals without the control algorithm's knowledge. The control gains were chosen as $K = \text{diag}(2, 2)$.

\section{Results And Discussion}
\subsection{Metric Trajectory Estimation}
The ATE results across all the 10 sequences are given in \cref{tab:atecompare} along with the path length and duration of the sequences. The proposed $\Phi$-constraint achieves lower ATE than VINS-Mono and ROVIO in all but two sequences. In those two sequences, it remains competitive. The $\tau$-constraint achieves a lower ATE than VINS-Mono in six out of nine comparable sequences and a lower ATE than ROVIO in 5 out of 10 comparable sequences. ATE averaged over all sequences was 5.4 cm for the $\Phi$-constraint, 8.5 cm for the $\tau$-onstraint, 16.9 cm for ROVIO, and 2.8 cm for AprilTag 3. The average ATE for VINS-Mono, averaged over all sequences except sequence 5, was 12.2 cm. The VINS-Mono result is omitted for sequence 5 because its estimate diverged.

It is no surprise that the AprilTag 3 method routinely achieved the best ATE. This is because the AprilTag system uses the known size of the visual fiducial to estimate depth. Regardless, the $\tau$ and $\Phi$-constraints perform comparably to the AprilTag 3 method in some sequences. This is particularly noticeable in Sequence 9. In \cref{fig:results_banner} the instantaneous $l_2$ error of each method in Sequence 9 is plotted for comparison.


While the ATE errors are promising, they do not indicate that our method is better than existing VIO methods. Such a claim would require developing a full VIO stack around the $\tau$ or $\Phi$-constraint and comparisons on existing datasets.

Our Python implementation achieved 6.5 $\times$ realtime or 588 frames per second (fps). VINS-Mono's C++ implementation ran at 0.26 $\times$ realtime or 23.6 fps. ROVIO's C++ implementation ran at 1.05 $\times$ realtime or 94.5 fps.

\subsection{Closed Loop Stability Invariance Property}
As shown in Fig. \ref{fig:djiexperimentsfull}, all achieved trajectories are similar, and approach the target, except for the case where the actuator gain was doubled and measured linear acceleration was used in the $\Phi$-constraint. This was expected. Indeed, in this case, the poles and zeros of the closed loop system are dramatically shifted because the control gain matrix is effectively doubled. As a result, the robot began to oscillate around its stopping point as is typical of a ``poorly tuned'' controller. However, the control scheme using efference copies had no such limitation, as predicted by Corollary \ref{thm:invariance}.

\section{Conclusion and Future Work}
In our work, we developed two novel constraints called the $\tau$ and $\Phi$-constraint which allow a moving camera to estimate depth using a small part of the image. Applying these constraints to trajectory estimation achieved better results while being orders of magnitude faster than some state-of-the-art VIO approaches. 
Further, we presented a method to perform closed-loop control with the constraints while using efference copies which is invariant to scaling of the control signal. We will talk about some future directions next.

Both constraints require that there is acceleration in order to measure distance.
In practice, we found accelerations with approximately 2 $\text{m}/\text{s}^2$ of power were necessary to get good measurements. However, when using efference copies only a small nominal acceleration was required for reasonable performance. Further theoretical analysis would be useful for gaining more insight into this behaviour.

VIO methods commonly estimate IMU biases and so it would be interesting to add such terms to our constraints. Similarly, it is of interest to extend \cref{thm:invariance} to account for a transfer function relating control effort and acceleration.


For our method to be deployable as a full VIO system it would need to be extended to fixate on multiple patches and actively switch between them.
Based on the significant speed up, and competitive accuracy presented in our preliminary results, we believe that further development of the $\tau$ and $\Phi$ constraint, in theory and practice, is a promising direction for VIO, VI-SLAM, active perception, and robotics.

\bibliographystyle{IEEEtran}
\bibliography{IEEEabrv, ms_amde_arxiv_v3.bib}

\end{document}


\title{Supplementary Material - TTCDist: Fast Distance Estimation From an Active Monocular Camera Using Time-to-Contact}

\maketitle

\section{Completion of Proof of Theorem 3.1}

For brevity, the manuscript left out the proof that the $\Phi$-constraint allows estimating initial depth $Z_0$ and initial velocity $\dot{\vect{X}}_0$ given measurements in an interval $t \in [0, T]$ if and only if $\exists \ t$ between the two times $ \text{s.t.} \  \ddot{\vect{X}} \neq 0$. In what follows we assume $\vect{X}(t)$ is sufficiently differentiable.

\begin{proof}
Recall the $\Phi$-constraint is

\begin{equation}\tag{$\Phi$-constraint}
\left(\Phi(t) - I \right) \begin{bmatrix}0 \\ 0 \\ Z_0\end{bmatrix} - t \dot{\vect{X}}_0 = \mathscr{J}\{\ddot{\vect{X}}\}(t).
\end{equation}

These three constraints are linear in $Z_0$ and $\vect{X}_0$. Thus, solving for these quantities using measurements over an interval $t \in [0, T]$ is not well posed if and only if $\Phi(t) - I$ and $-t$ are linearly dependent over the interval.

Consider that
\begin{equation}
\begin{split}
&\vect{X}(t) = \Phi(t)\vect{X}_0 \\
\iff &\vect{X}(t) - \vect{X}_0 = (\Phi(t) - I) \vect{X}_0\\
\iff &\vect{X}(t) - \vect{X}_0 = (\Phi(t) - I) \begin{bmatrix}0 \\ 0 \\ Z_0\end{bmatrix} \\
\iff &\frac{\vect{X}(t) - \vect{X}_0}{Z_0} = (\Phi(t) - I) \begin{bmatrix}0 \\ 0 \\ 1\end{bmatrix}.
\end{split}
\end{equation}

Therefore it is equivalent to determine when $ (\vect{X}(t) - \vect{X}_0) / Z_0$ and $-t$ are linearly dependent. Assuming the two functions are equal up to a scalar $\alpha \in \mathbb{R}$ reveals

\begin{equation}
\begin{split}
     &\frac{\vect{X}(t) - \vect{X}_0}{Z_0} = \alpha (-t)\\
\iff &\vect{X}(t) = \vect{X}_0 + \alpha (-t) Z_0\\
\iff &\ddot{\vect{X}}(t) = 0 .
\end{split}
\end{equation}

Thus, the functions are linearly independent, and so $Z_0$ and $\vect{X}_0$ can be determined, if and only if $\exists \ t \in [0, T] \ \text{s.t.} \ \ddot{\vect{X}}(t) \neq 0$.
\end{proof}

\section{Completion of Proof of Theorem 3.2}

For brevity, the manuscript left out the proof that the $\tau$-constraint allows estimating initial depth $Z_0$ given measurements in an interval $t \in [0, T]$ and if and only if $\exists \ t$ between the two times $ \text{s.t.} \  \ddot{\vect{X}} \neq 0$.

\begin{proof}
Recall the $\tau$-constraint is

\begin{equation}\tag{$\tau$-constraint}
\underbrace{\left(\Phi(t) - I - t\begin{bmatrix} 0 & 0 & \vect{F}(0)\end{bmatrix}\right)}_{E(t) \coloneqq}\begin{bmatrix}0 \\ 0 \\ Z_0\end{bmatrix} = \mathscr{J}\{\ddot{\vect{X}}\}(t).
\end{equation}

The constraint is linear in $Z_0$ thus solving for $Z_0$ quantities using measurements over an interval $t \in [0, T]$ is not well posed if and only if $E(t) = 0 \ \forall \ t \in [0, T]$. Substituting the previous expression for $\Phi(t) - I$ and the definition of $F(0) = \dot{\vect{X}}_0 / Z_0$ reveals

\begin{equation}
\begin{split}
     & \Phi(t) - I - t\begin{bmatrix} 0 & 0 & \vect{F}(0)\end{bmatrix} = 0\\
\iff & \frac{\vect{X}(t) - \vect{X}_0}{Z_0} - t\frac{\dot{\vect{X}}_0}{Z_0} = 0\\
\iff & \vect{X}(t) = \vect{X}_0 + t\dot{\vect{X}}_0\\
\iff & \ddot{\vect{X}}(t) = 0
\end{split}
\end{equation}

Therefore, the problem is well-posed (meaning $Z_0$ can be determined) if and only if $\exists \ t \in [0, T] \ \text{s.t.} \ \ddot{\vect{X}}(t) \neq 0$.
\end{proof}

\section{Proof that Eq. (7) and (8) are well posed}
Equations (7) and (8) setup a linear least squares problem to determine $Z_0$, $\dot{Z}_0$ and $g_Z$ from $\Phi(t)-I$ and $a^m_Z$. It is stated the problem is the same for the X or Y axis up to a change of subscripts and the initial velocity and claimed without proof that the problems are well-posed if and only if, $a^m$ is not constant for the entire time interval. The proof follows.

\begin{proof}

For Eq. (7), the linear relation being optimized (based on the $\Phi$-constraint) and considered along each axis at once is

\begin{equation}
\begin{split}
& \left(\Phi(t) - 1\right) \begin{bmatrix}0 \\ 0 \\ Z_0\end{bmatrix} - t \dot{\vect{X}}_0 +  \mathscr{J}\{\vect{a}^m + \vect{g}\}(t) = 0\\
\iff &\frac{\vect{X}(t) - \vect{X}_0}{Z_0} Z_0 - t \dot{\vect{X}}_0 + \vect{g} \frac{t^2}{2} + \mathscr{J}\{\vect{a}^m\}(t) = 0.
\end{split}
\end{equation}

Similarly to Theorem 3.1 and 3.2, the problem is not well posed if and only if $(\vect{X}(t) - \vect{X}_0) / Z_0$, $-t$, and $t^2/2$ are linearly dependent $\forall \ t \in [0, T]$. Assuming $\exists \ \alpha, \beta \in \mathbb{R}$ s.t. a weighted linear combination equals zero reveals


\begin{equation}
\begin{split}
    &\frac{\vect{X}(t) - \vect{X}_0}{Z_0} - \alpha t + \beta \frac{t^2}{2} = 0\\
\iff &\vect{X}(t) = \vect{X}_0 + \alpha Z_0 t - \beta Z_0 \frac{t^2}{2} = 0\\
\iff &\frac{d}{dt} \ddot{\vect{X}}(t) = 0.
\end{split}
\end{equation}

The proof and result for Eq. (9) (based on the $\tau$-constraint) is identical with the exception that $\alpha Z_0 t$ in the second to last line becomes $\alpha \dot{\vect{X}}_0 t$.

Thus, the problems in Eq. (8) and (9) are well-posed if and only if the derivative of acceleration, jerk, is non-zero at some time during the considered time interval.
\end{proof}

\iftrue

\section{Sample Images, Trajectory, and Error Plots for all Sequences}

Below, for each of the ten sequences, we show four sample images, $X$, $Y$, $Z$ trajectory estimates, and $l_2$ error with regard to the Vicon ground truth for every trajectory estimation method available. The plots are to give a sense of the qualitative properties of each algorithm. All trajectories are aligned to the vicon ground-truth by a constant rotation and translation.

When run on Sequence 5, VINS-Mono diverged, so its results are not plotted for that sequence.

The biggest qualitative flaw with the $\tau$-constraint's estimates are occasional ``spikes''. These are caused by occasional, large errors in the frequency-of-contact measurements that can happen as a result of the finite-difference scheme used. The $\Phi$ constraint does not have this issue.

\begin{figure*}
  \centering
  \begin{subfigure}{0.48\linewidth}
    \includegraphics[width=1.0\linewidth]{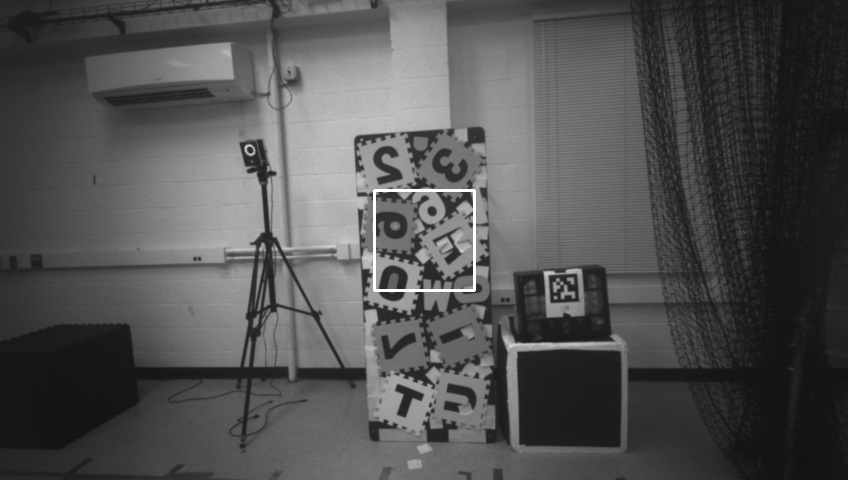}
  \end{subfigure}
  \begin{subfigure}{0.48\linewidth}
    \includegraphics[width=1.0\linewidth]{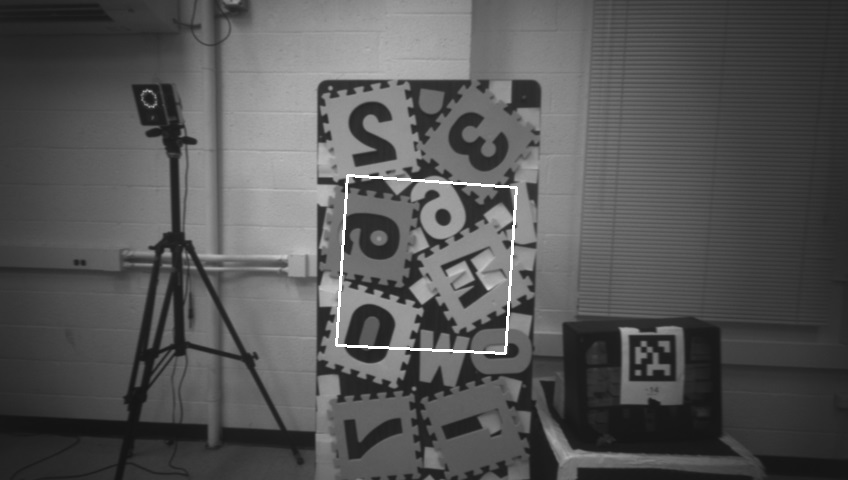}
  \end{subfigure}
  \begin{subfigure}{0.48\linewidth}
    \includegraphics[width=1.0\linewidth]{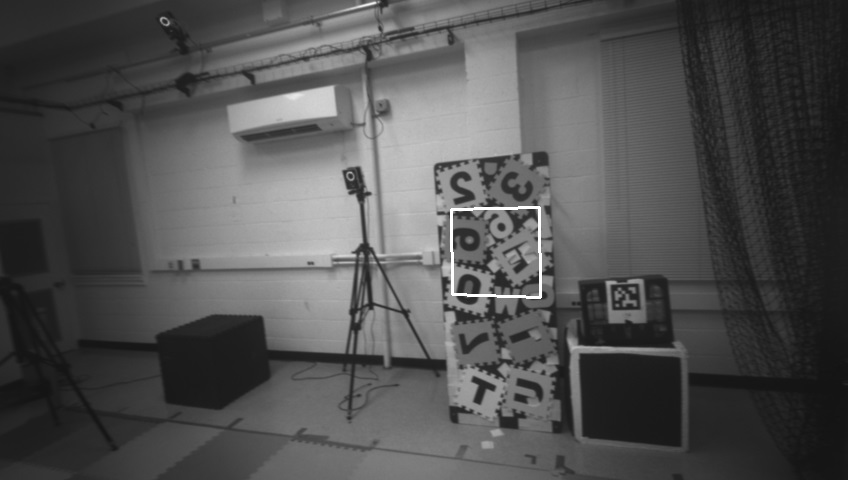}
  \end{subfigure}
  \begin{subfigure}{0.48\linewidth}
    \includegraphics[width=1.0\linewidth]{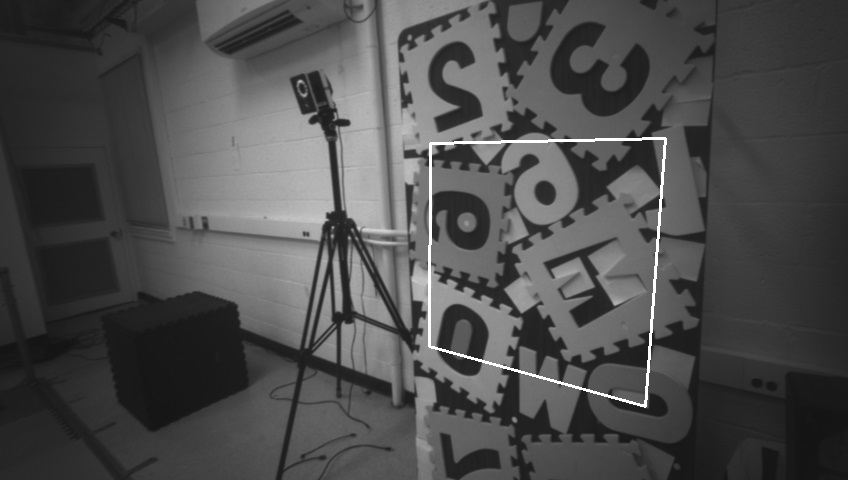}
  \end{subfigure}
  \caption{Four representative frames from Sequence 1. The tracked patch is indicated by the white box.} 
\label{fig:record_000000_frames}
\end{figure*}

\begin{figure*}
  \centering
  \begin{subfigure}{0.48\linewidth}
    \includegraphics[width=1.0\linewidth]{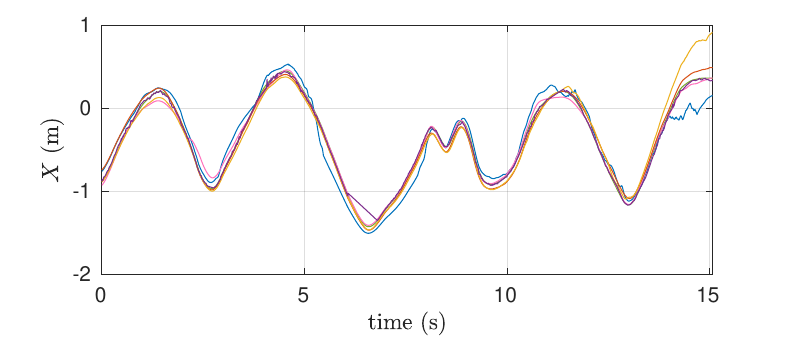}
  \end{subfigure}
  \begin{subfigure}{0.48\linewidth}
    \includegraphics[width=1.0\linewidth]{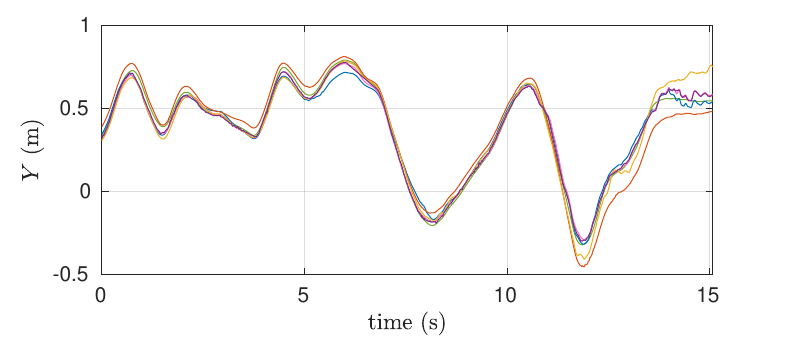}
  \end{subfigure}
  \begin{subfigure}{0.48\linewidth}
    \includegraphics[width=1.0\linewidth]{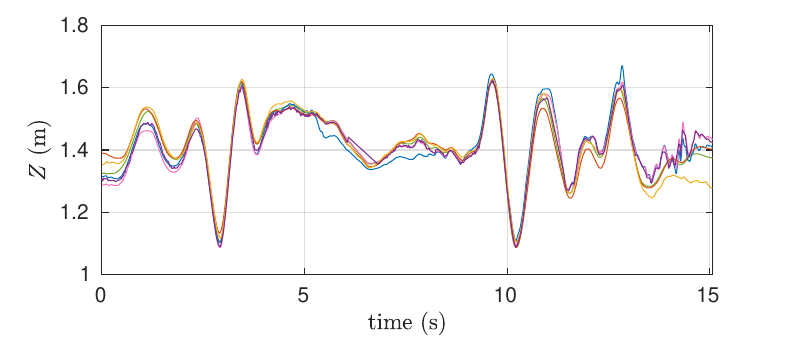}
  \end{subfigure}
  \begin{subfigure}{0.48\linewidth}
    \includegraphics[width=1.0\linewidth]{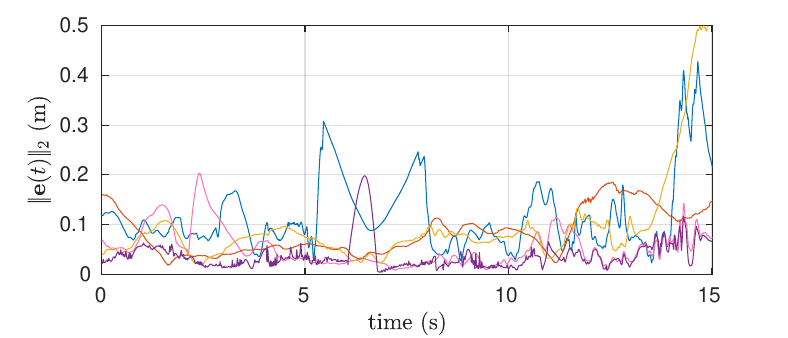}
  \end{subfigure}
  \begin{subfigure}{0.5\linewidth}
    \includegraphics[width=1.0\linewidth]{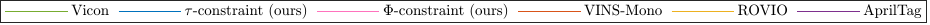}
  \end{subfigure}
  \caption{Sequence 1 estimated trajectories versus Vicon ground truth for all available estimation methods. All trajectories are aligned to the vicon by a rigid body transformation. (top-left) X axis trajectory, (top-right) Y axis trajectory, (bottom-left) Z axis trajectory, (bottom-right) $l_2$ error between estimated trajectories and Vicon ground truth.}
  \label{fig:record_000000_trajectory}
\end{figure*}

\begin{figure*}
  \centering
  \begin{subfigure}{0.48\linewidth}
    \includegraphics[width=1.0\linewidth]{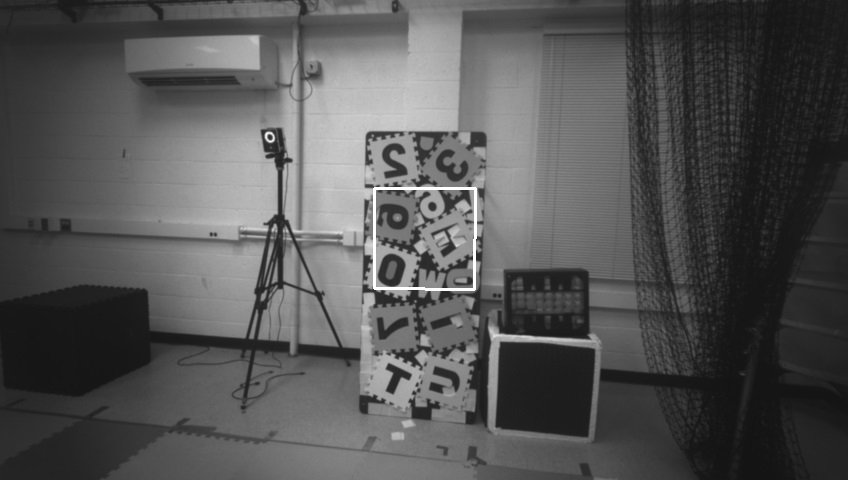}
  \end{subfigure}
  \begin{subfigure}{0.48\linewidth}
    \includegraphics[width=1.0\linewidth]{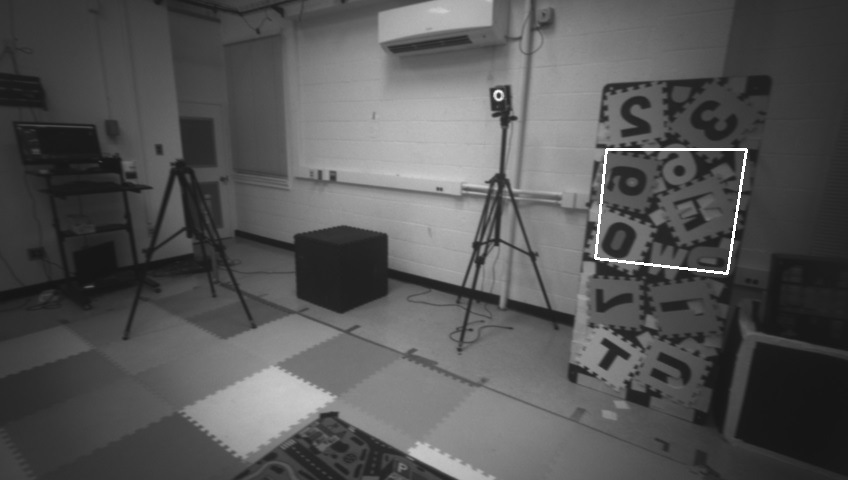}
  \end{subfigure}
  \begin{subfigure}{0.48\linewidth}
    \includegraphics[width=1.0\linewidth]{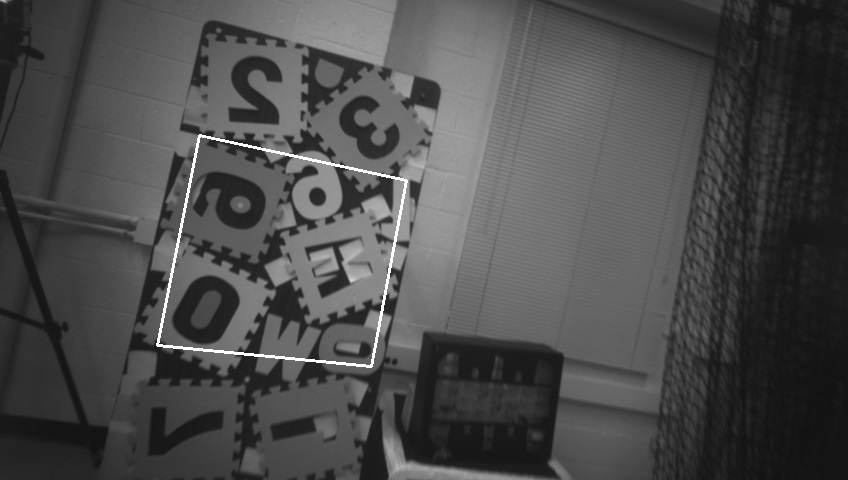}
  \end{subfigure}
  \begin{subfigure}{0.48\linewidth}
    \includegraphics[width=1.0\linewidth]{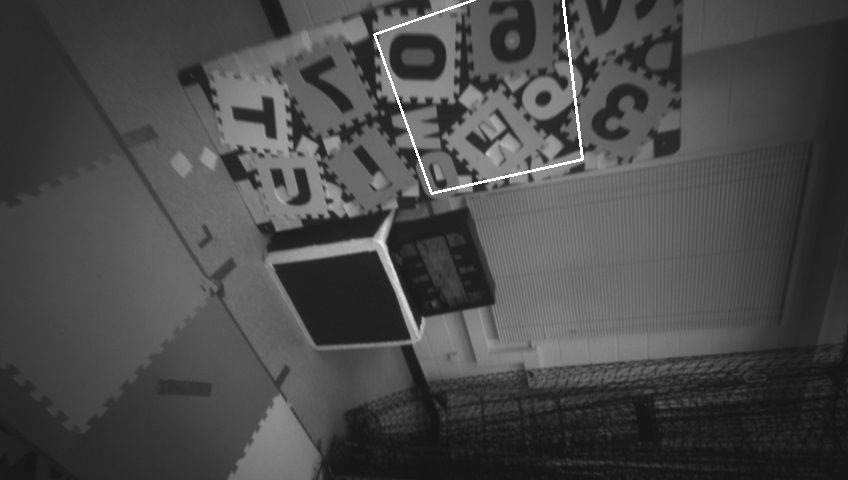}
  \end{subfigure}
  \caption{Four representative frames from Sequence 2. The tracked patch is indicated by the white box. This sequence does not contain an AprilTag.} 
\label{fig:record_000001_frames}
\end{figure*}

\begin{figure*}
  \centering
  \begin{subfigure}{0.48\linewidth}
    \includegraphics[width=1.0\linewidth]{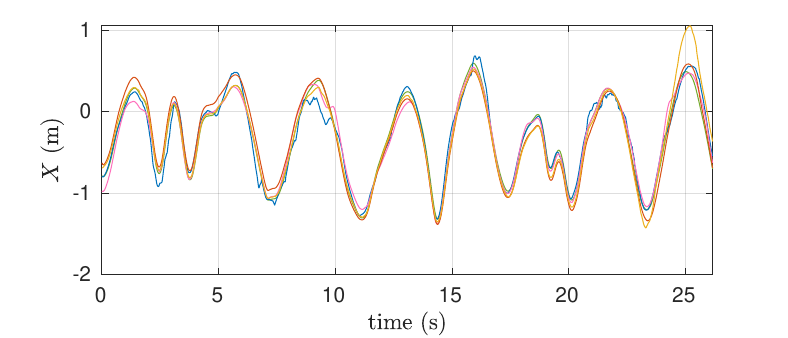}
  \end{subfigure}
  \begin{subfigure}{0.48\linewidth}
    \includegraphics[width=1.0\linewidth]{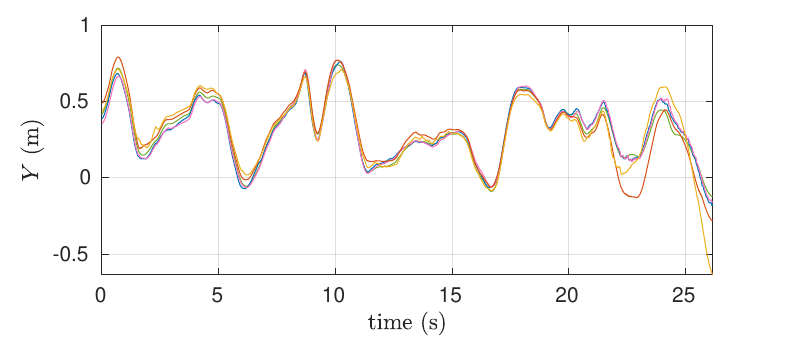}
  \end{subfigure}
  \begin{subfigure}{0.48\linewidth}
    \includegraphics[width=1.0\linewidth]{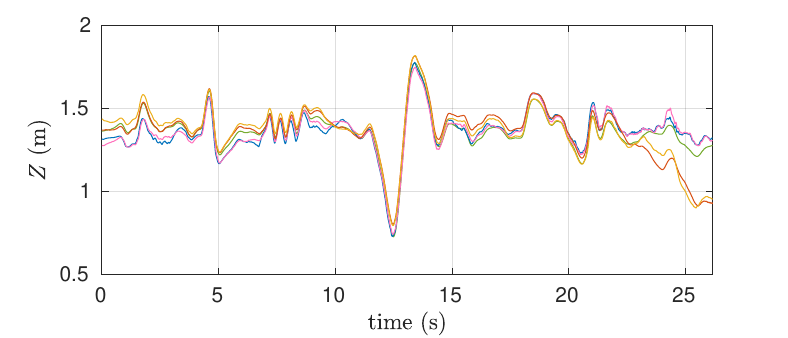}
  \end{subfigure}
  \begin{subfigure}{0.48\linewidth}
    \includegraphics[width=1.0\linewidth]{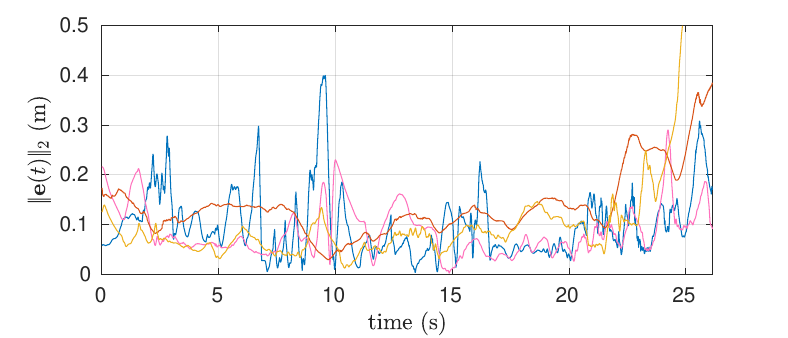}
  \end{subfigure}
  \begin{subfigure}{0.5\linewidth}
    \includegraphics[width=1.0\linewidth]{figures_supp/legend.pdf}
  \end{subfigure}
  \caption{Sequence 2 estimated trajectories versus Vicon ground truth for all available estimation methods. All trajectories are aligned to the vicon by a rigid body transformation. (top-left) X axis trajectory, (top-right) Y axis trajectory, (bottom-left) Z axis trajectory, (bottom-right) $l_2$ error between estimated trajectories and Vicon ground truth.}
  \label{fig:record_000001_trajectory}
\end{figure*}

\begin{figure*}
  \centering
  \begin{subfigure}{0.48\linewidth}
    \includegraphics[width=1.0\linewidth]{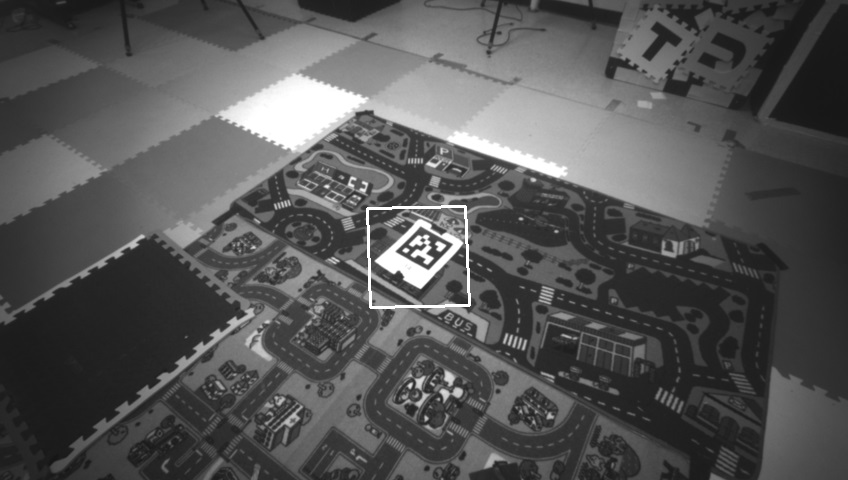}
  \end{subfigure}
  \begin{subfigure}{0.48\linewidth}
    \includegraphics[width=1.0\linewidth]{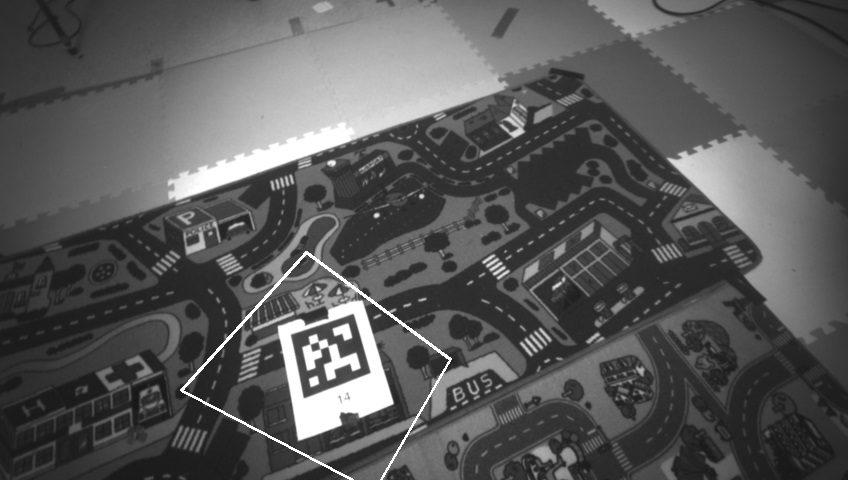}
  \end{subfigure}
  \begin{subfigure}{0.48\linewidth}
    \includegraphics[width=1.0\linewidth]{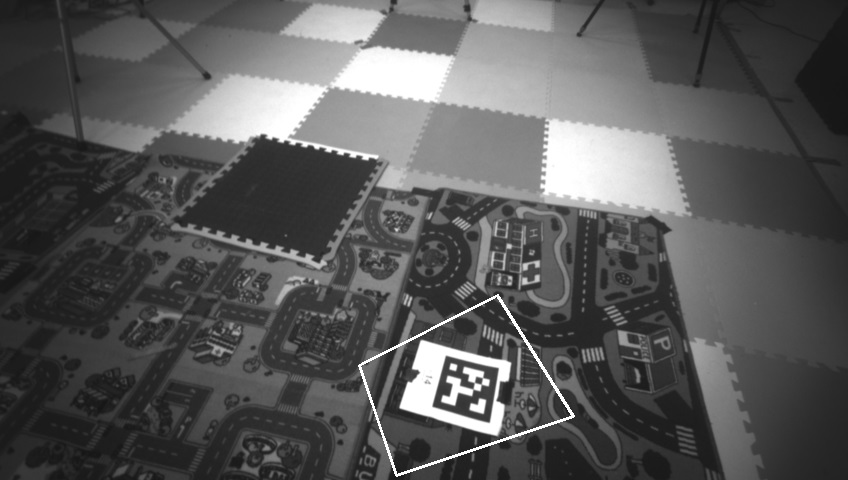}
  \end{subfigure}
  \begin{subfigure}{0.48\linewidth}
    \includegraphics[width=1.0\linewidth]{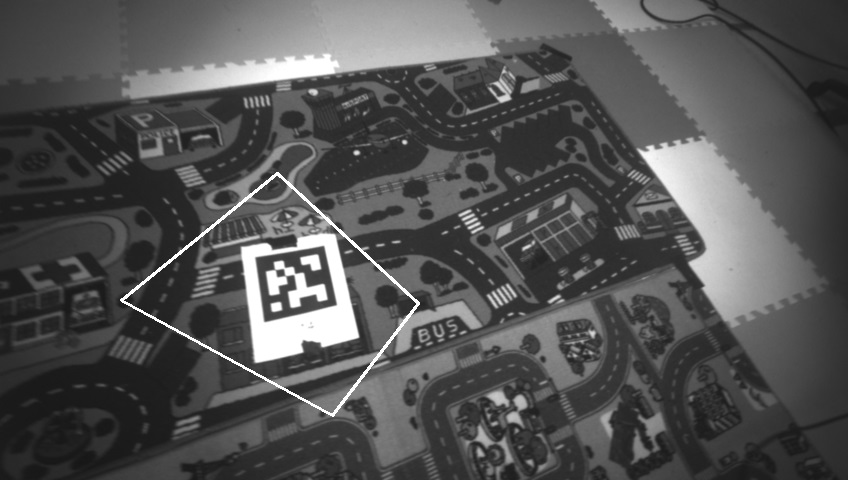}
  \end{subfigure}
  \caption{Four representative frames from Sequence 3. The tracked patch is indicated by the white box.} 
\label{fig:record_000002_frames}
\end{figure*}

\begin{figure*}
  \centering
  \begin{subfigure}{0.48\linewidth}
    \includegraphics[width=1.0\linewidth]{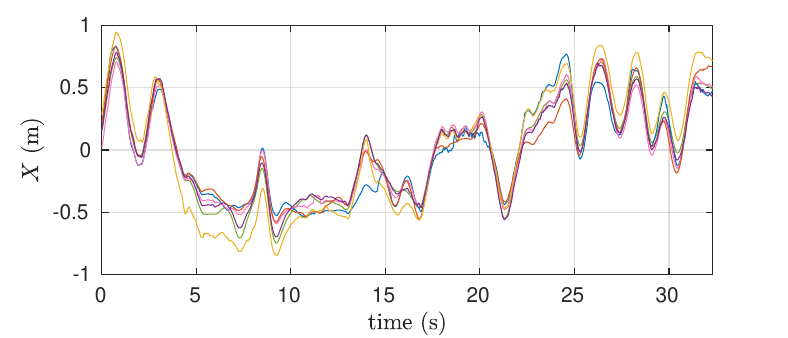}
  \end{subfigure}
  \begin{subfigure}{0.48\linewidth}
    \includegraphics[width=1.0\linewidth]{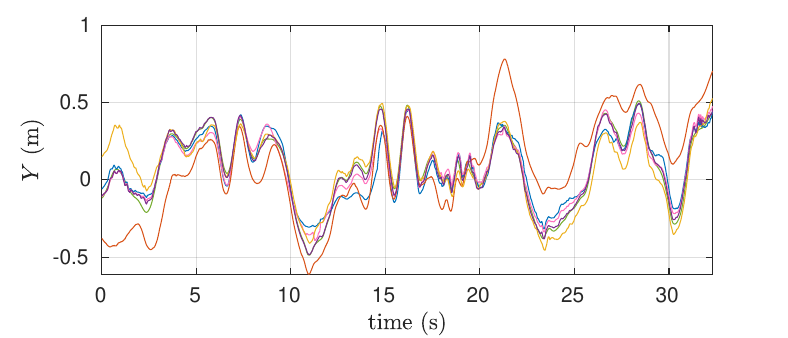}
  \end{subfigure}
  \begin{subfigure}{0.48\linewidth}
    \includegraphics[width=1.0\linewidth]{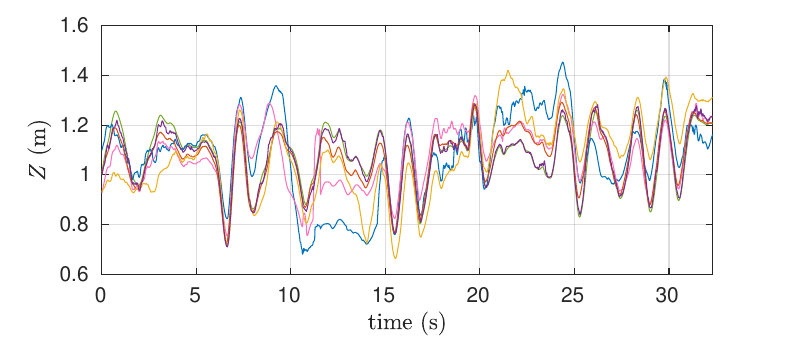}
  \end{subfigure}
  \begin{subfigure}{0.48\linewidth}
    \includegraphics[width=1.0\linewidth]{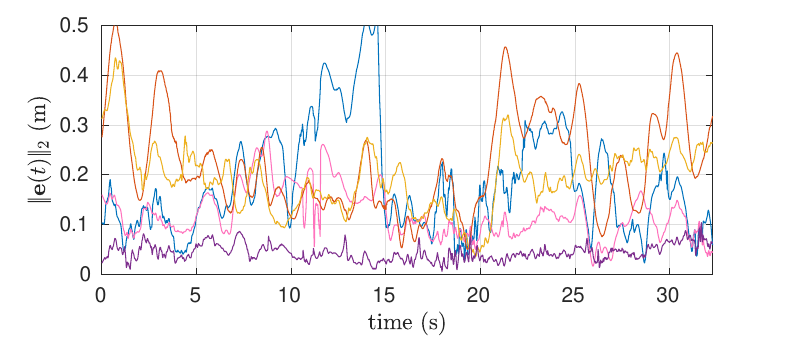}
  \end{subfigure}
  \begin{subfigure}{0.5\linewidth}
    \includegraphics[width=1.0\linewidth]{figures_supp/legend.pdf}
  \end{subfigure}
  \caption{Sequence 3 estimated trajectories versus Vicon ground truth for all available estimation methods. All trajectories are aligned to the vicon by a rigid body transformation.  (top-left) X axis trajectory, (top-right) Y axis trajectory, (bottom-left) Z axis trajectory, (bottom-right) $l_2$ error between estimated trajectories and Vicon ground truth.}
  \label{fig:record_000002_trajectory}
\end{figure*}

\begin{figure*}
  \centering
  \begin{subfigure}{0.48\linewidth}
    \includegraphics[width=1.0\linewidth]{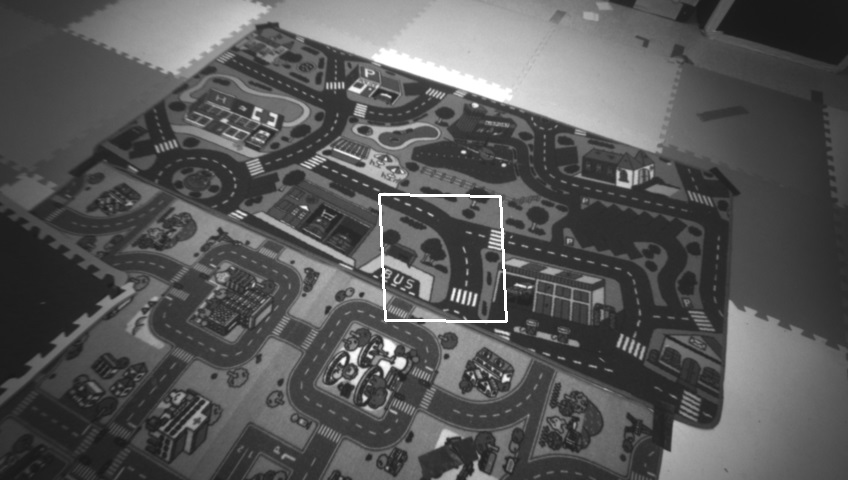}
  \end{subfigure}
  \begin{subfigure}{0.48\linewidth}
    \includegraphics[width=1.0\linewidth]{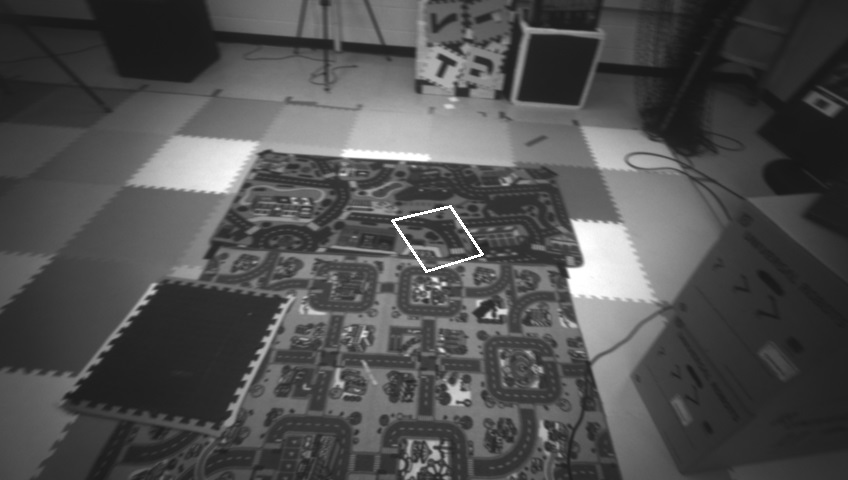}
  \end{subfigure}
  \begin{subfigure}{0.48\linewidth}
    \includegraphics[width=1.0\linewidth]{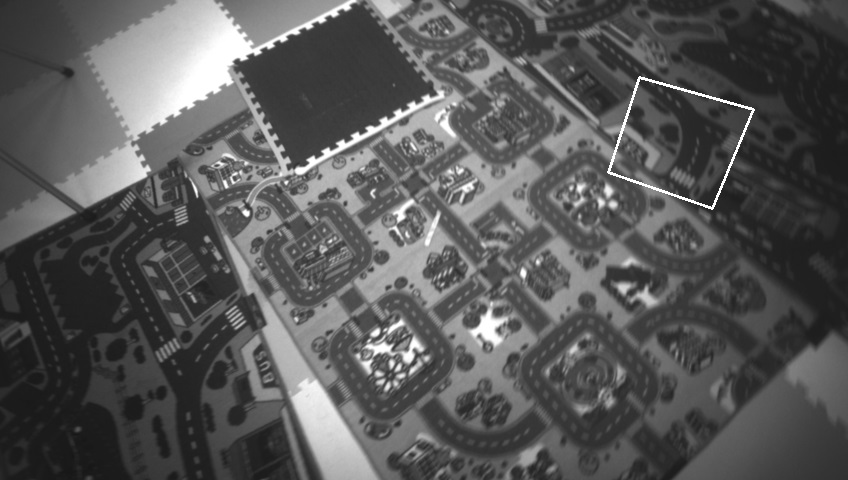}
  \end{subfigure}
  \begin{subfigure}{0.48\linewidth}
    \includegraphics[width=1.0\linewidth]{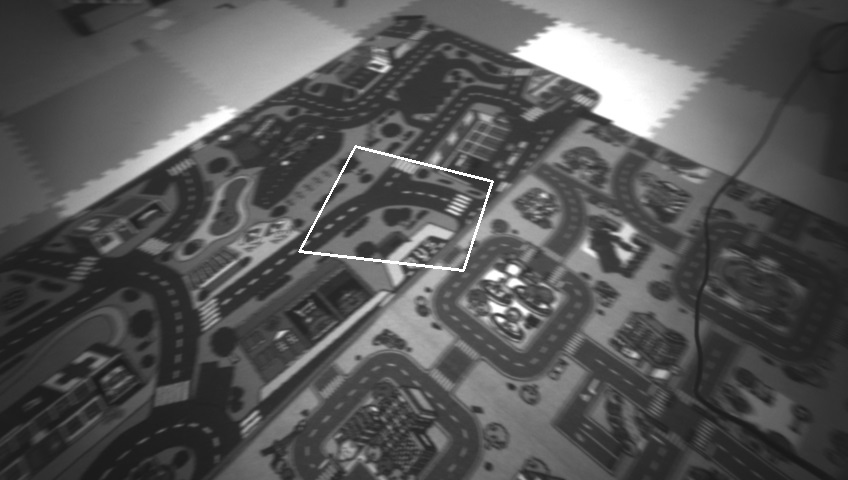}
  \end{subfigure}
  \caption{Four representative frames from Sequence 4. The tracked patch is indicated by the white box. This sequence does not contain an AprilTag.} 
\label{fig:record_000003_frames}
\end{figure*}

\begin{figure*}
  \centering
  \begin{subfigure}{0.48\linewidth}
    \includegraphics[width=1.0\linewidth]{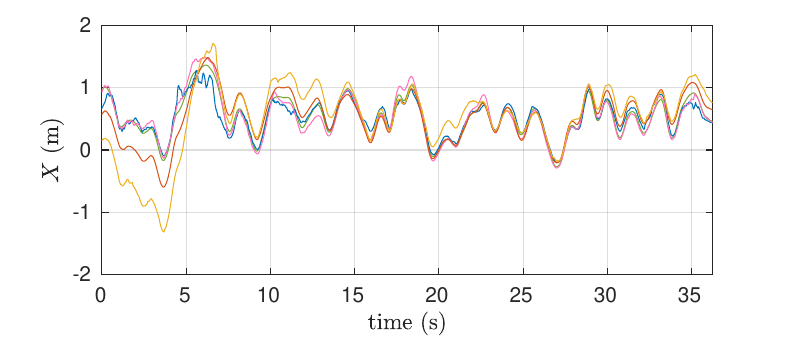}
  \end{subfigure}
  \begin{subfigure}{0.48\linewidth}
    \includegraphics[width=1.0\linewidth]{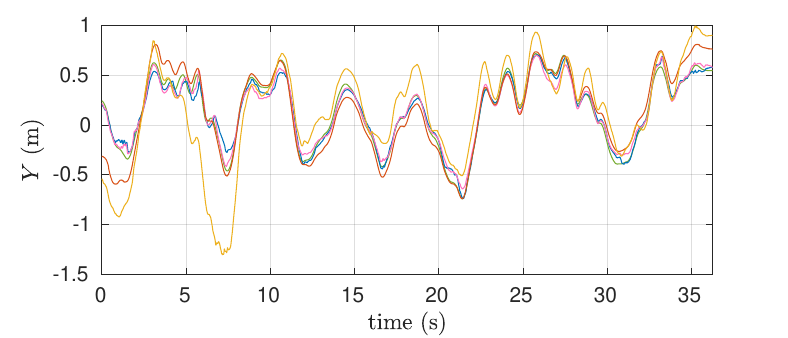}
  \end{subfigure}
  \begin{subfigure}{0.48\linewidth}
    \includegraphics[width=1.0\linewidth]{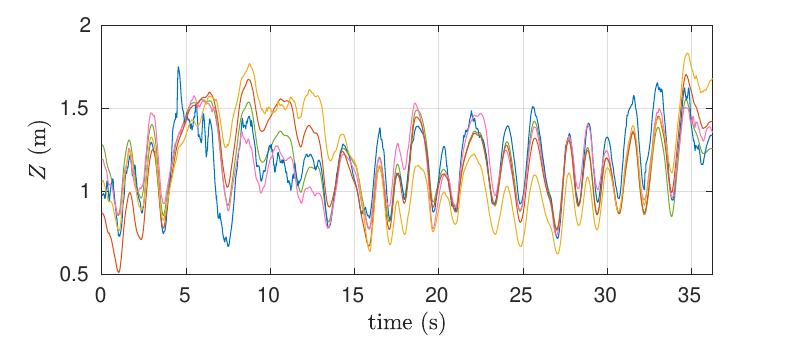}
  \end{subfigure}
  \begin{subfigure}{0.48\linewidth}
    \includegraphics[width=1.0\linewidth]{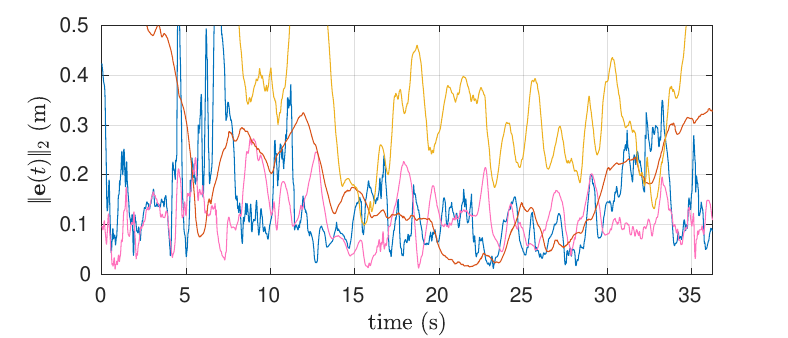}
  \end{subfigure}
  \begin{subfigure}{0.5\linewidth}
    \includegraphics[width=1.0\linewidth]{figures_supp/legend.pdf}
  \end{subfigure}
  \caption{Sequence 4 estimated trajectories versus Vicon ground truth for all available estimation methods.  All trajectories are aligned to the vicon by a rigid body transformation.  (top-left) X axis trajectory, (top-right) Y axis trajectory, (bottom-left) Z axis trajectory, (bottom-right) $l_2$ error between estimated trajectories and Vicon ground truth.}
  \label{fig:record_000003_trajectory}
\end{figure*}

\begin{figure*}
  \centering
  \begin{subfigure}{0.48\linewidth}
    \includegraphics[width=1.0\linewidth]{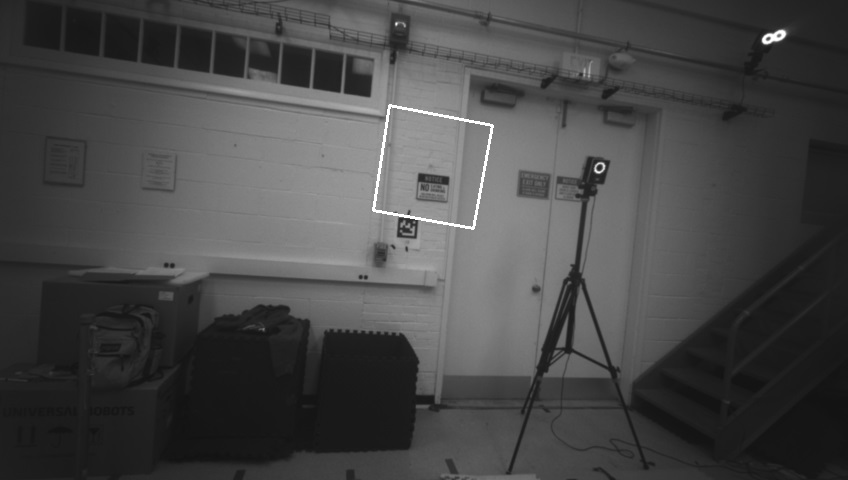}
  \end{subfigure}
  \begin{subfigure}{0.48\linewidth}
    \includegraphics[width=1.0\linewidth]{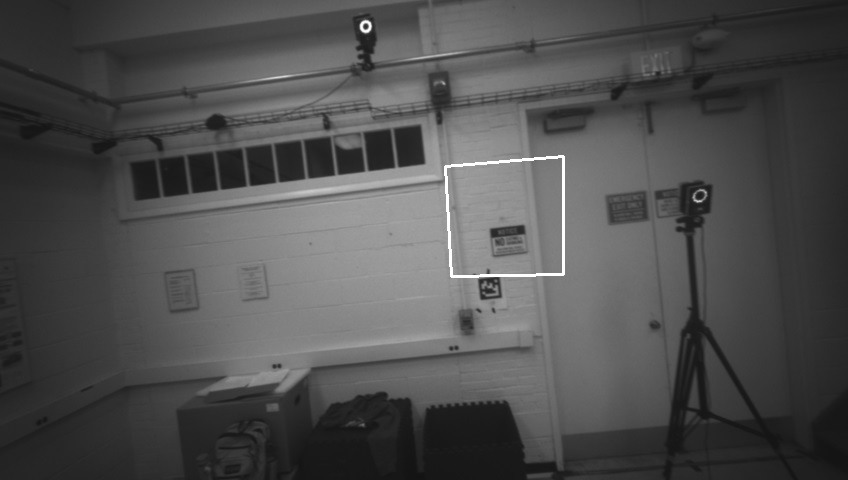}
  \end{subfigure}
  \begin{subfigure}{0.48\linewidth}
    \includegraphics[width=1.0\linewidth]{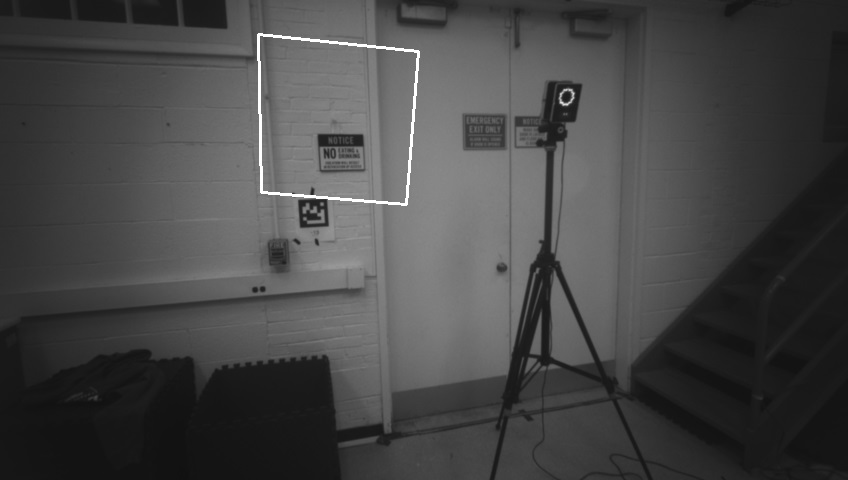}
  \end{subfigure}
  \begin{subfigure}{0.48\linewidth}
    \includegraphics[width=1.0\linewidth]{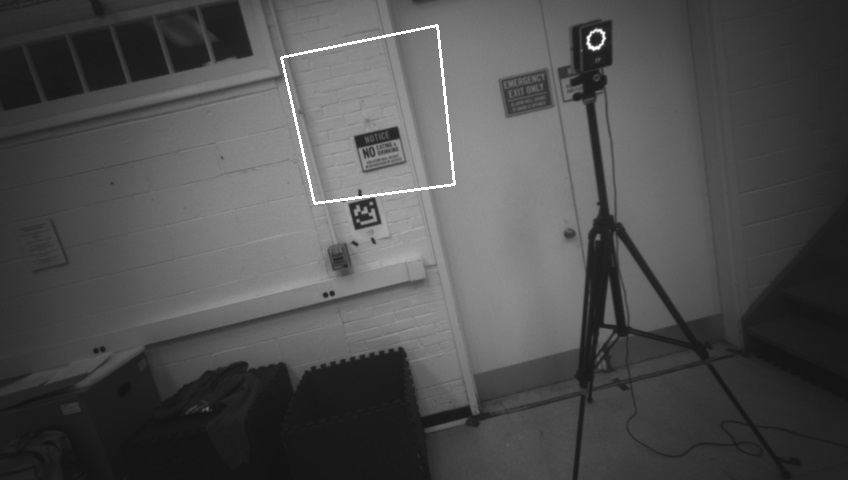}
  \end{subfigure}
  \caption{Four representative frames from Sequence 5. The tracked patch is indicated by the white box.} 
\label{fig:record_000004_frames}
\end{figure*}

\begin{figure*}
  \centering
  \begin{subfigure}{0.48\linewidth}
    \includegraphics[width=1.0\linewidth]{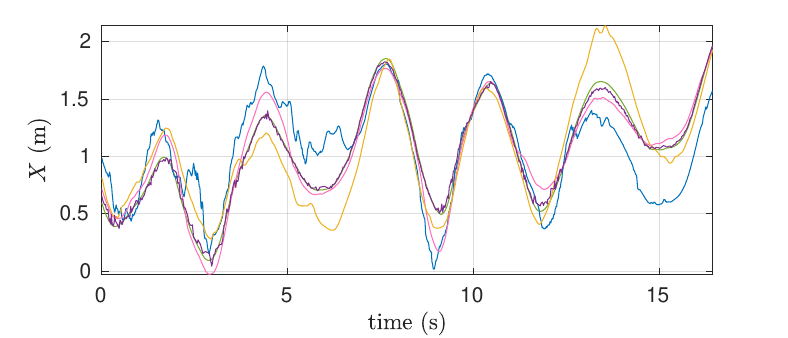}
  \end{subfigure}
  \begin{subfigure}{0.48\linewidth}
    \includegraphics[width=1.0\linewidth]{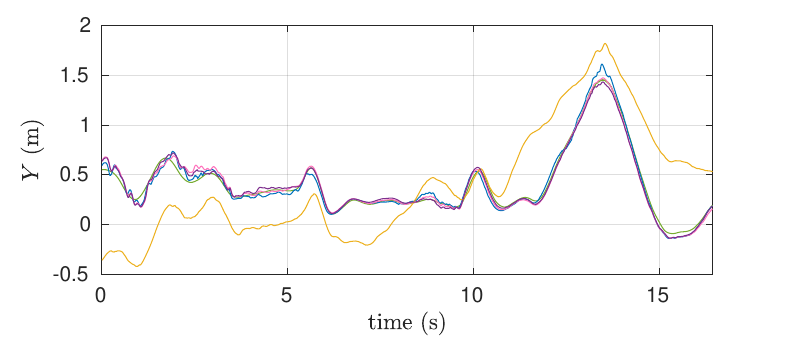}
  \end{subfigure}
  \begin{subfigure}{0.48\linewidth}
    \includegraphics[width=1.0\linewidth]{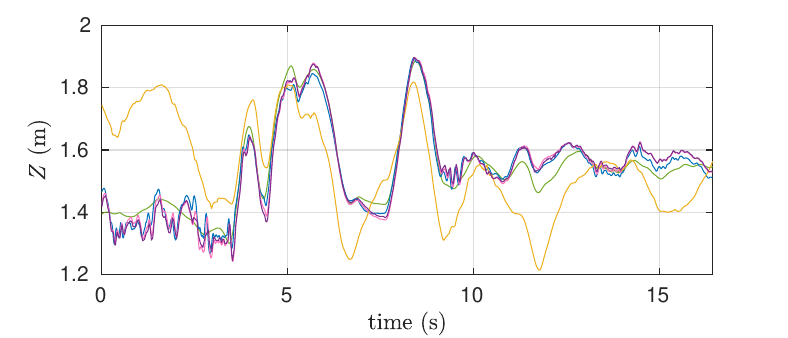}
  \end{subfigure}
  \begin{subfigure}{0.48\linewidth}
    \includegraphics[width=1.0\linewidth]{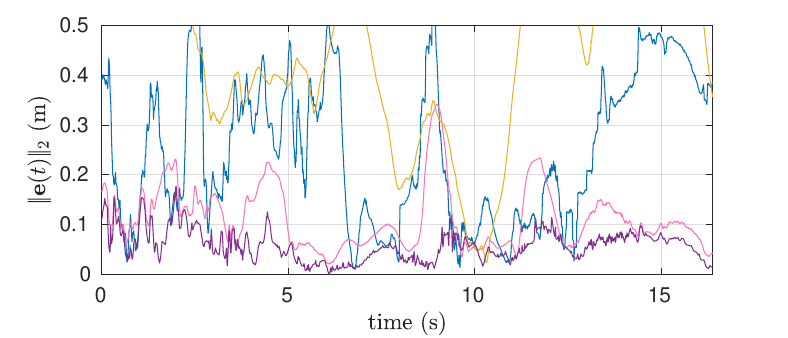}
  \end{subfigure}
  \begin{subfigure}{0.5\linewidth}
    \includegraphics[width=1.0\linewidth]{figures_supp/legend.pdf}
  \end{subfigure}
  \caption{Sequence 5 estimated trajectories versus Vicon ground truth for all available estimation methods. VINS-Mono is left out because it diverged.  All trajectories are aligned to the vicon by a rigid body transformation.  (top-left) X axis trajectory, (top-right) Y axis trajectory, (bottom-left) Z axis trajectory, (bottom-right) $l_2$ error between estimated trajectories and Vicon ground truth.}
  \label{fig:record_000004_trajectory}
\end{figure*}

\begin{figure*}
  \centering
  \begin{subfigure}{0.48\linewidth}
    \includegraphics[width=1.0\linewidth]{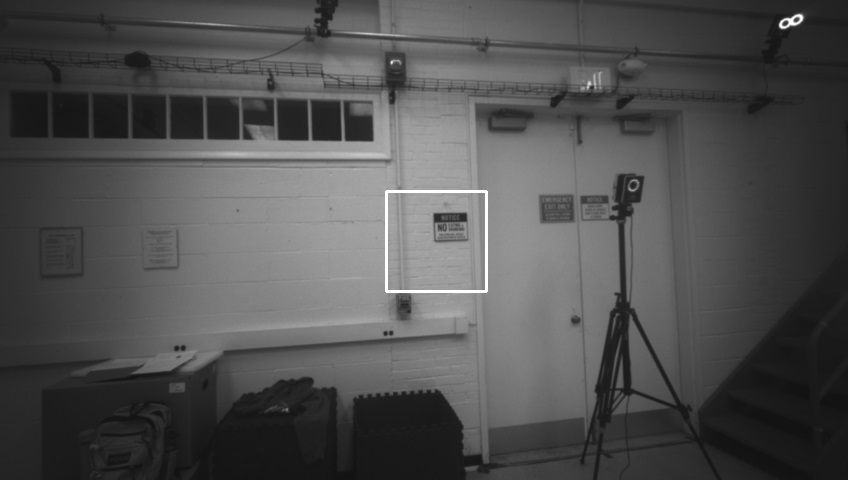}
  \end{subfigure}
  \begin{subfigure}{0.48\linewidth}
    \includegraphics[width=1.0\linewidth]{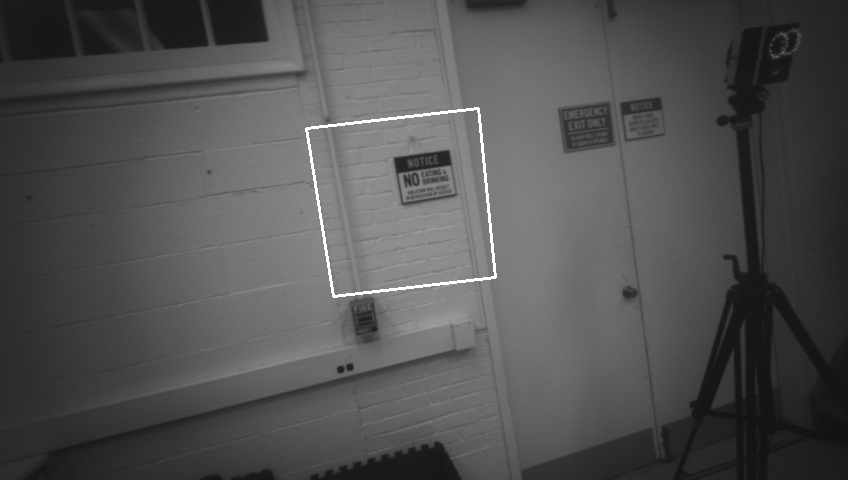}
  \end{subfigure}
  \begin{subfigure}{0.48\linewidth}
    \includegraphics[width=1.0\linewidth]{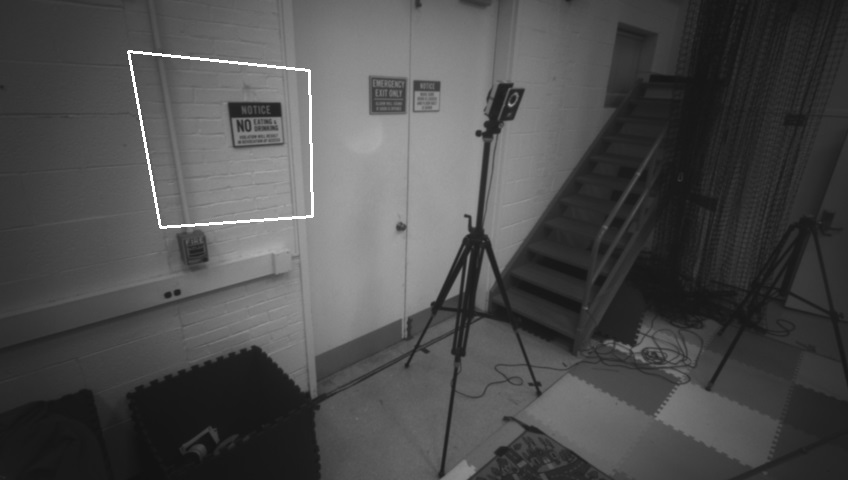}
  \end{subfigure}
  \begin{subfigure}{0.48\linewidth}
    \includegraphics[width=1.0\linewidth]{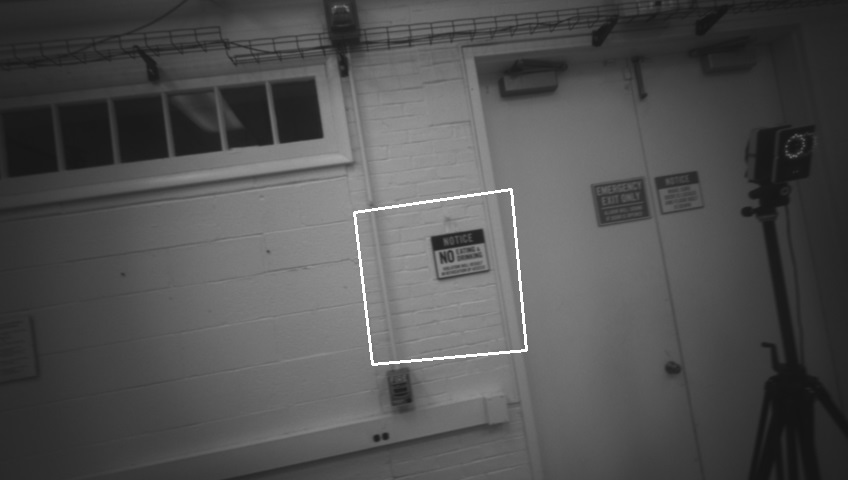}
  \end{subfigure}
  \caption{Four representative frames from Sequence 6. The tracked patch is indicated by the white box. This sequence does not contain an AprilTag.} 
\label{fig:record_000005_frames}
\end{figure*}

\begin{figure*}
  \centering
  \begin{subfigure}{0.48\linewidth}
    \includegraphics[width=1.0\linewidth]{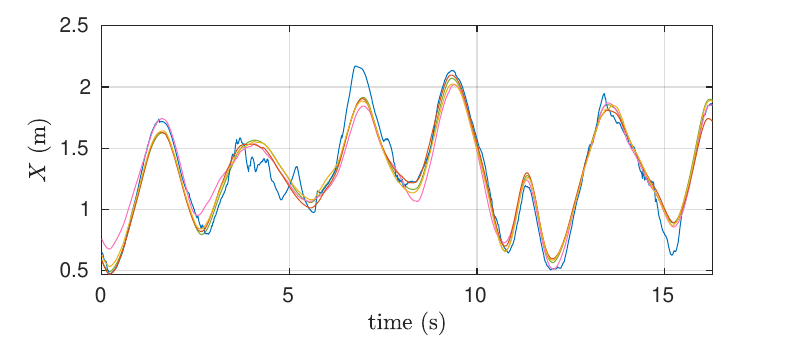}
  \end{subfigure}
  \begin{subfigure}{0.48\linewidth}
    \includegraphics[width=1.0\linewidth]{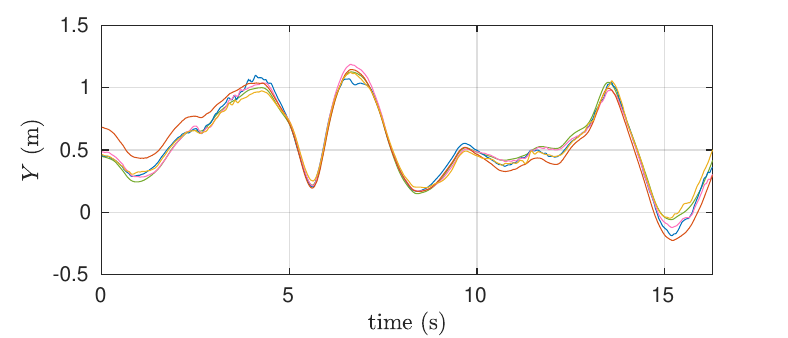}
  \end{subfigure}
  \begin{subfigure}{0.48\linewidth}
    \includegraphics[width=1.0\linewidth]{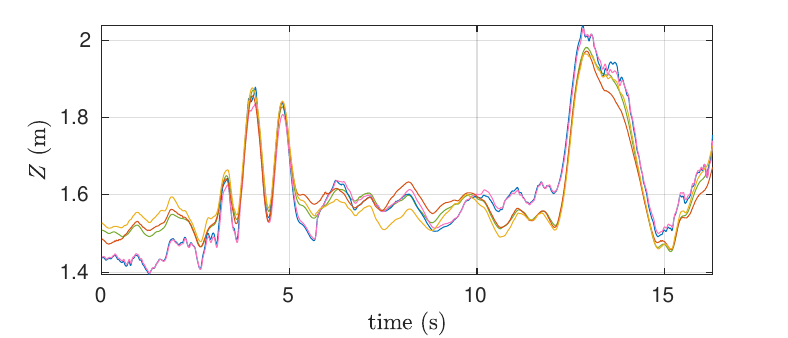}
  \end{subfigure}
  \begin{subfigure}{0.48\linewidth}
    \includegraphics[width=1.0\linewidth]{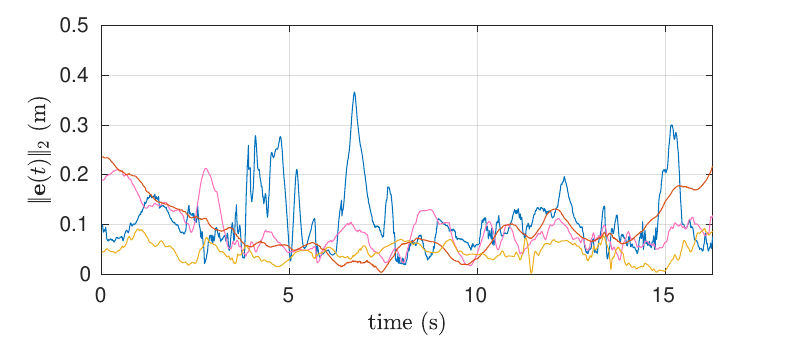}
  \end{subfigure}
  \begin{subfigure}{0.5\linewidth}
    \includegraphics[width=1.0\linewidth]{figures_supp/legend.pdf}
  \end{subfigure}
  \caption{Sequence 6 estimated trajectories versus Vicon ground truth for all available estimation methods.  All trajectories are aligned to the vicon by a rigid body transformation.  (top-left) X axis trajectory, (top-right) Y axis trajectory, (bottom-left) Z axis trajectory, (bottom-right) $l_2$ error between estimated trajectories and Vicon ground truth.}
  \label{fig:record_000005_trajectory}
\end{figure*}

\begin{figure*}
  \centering
  \begin{subfigure}{0.48\linewidth}
    \includegraphics[width=1.0\linewidth]{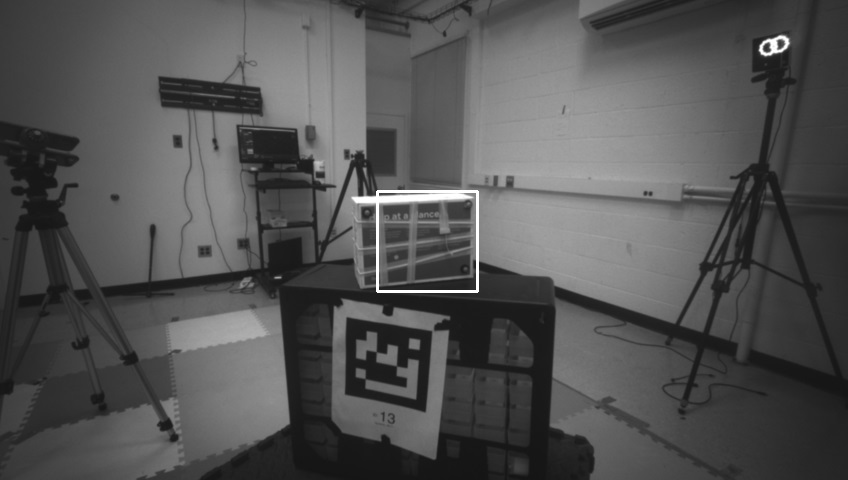}
  \end{subfigure}
  \begin{subfigure}{0.48\linewidth}
    \includegraphics[width=1.0\linewidth]{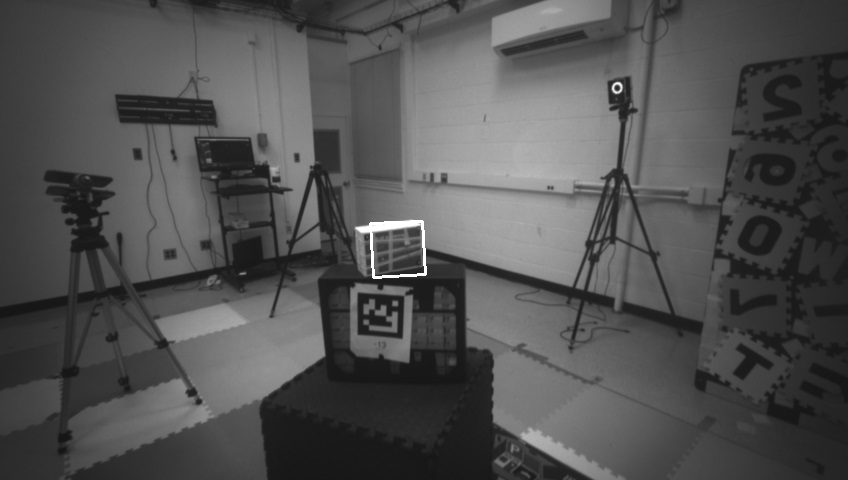}
  \end{subfigure}
  \begin{subfigure}{0.48\linewidth}
    \includegraphics[width=1.0\linewidth]{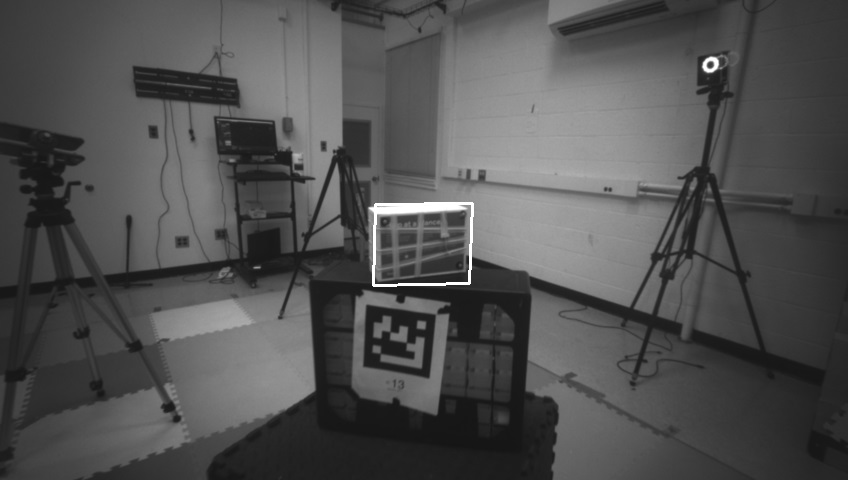}
  \end{subfigure}
  \begin{subfigure}{0.48\linewidth}
    \includegraphics[width=1.0\linewidth]{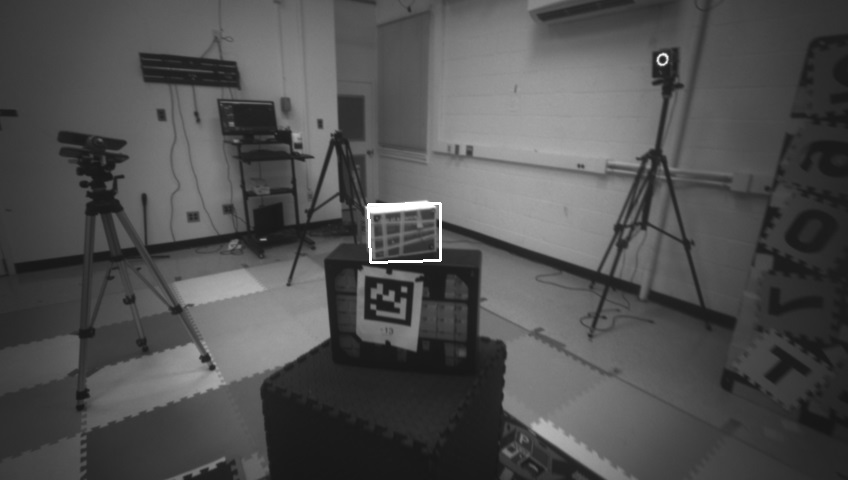}
  \end{subfigure}
  \caption{Four representative frames from Sequence 7. The tracked patch is indicated by the white box.} 
\label{fig:record_000006_frames}
\end{figure*}

\begin{figure*}
  \centering
  \begin{subfigure}{0.48\linewidth}
    \includegraphics[width=1.0\linewidth]{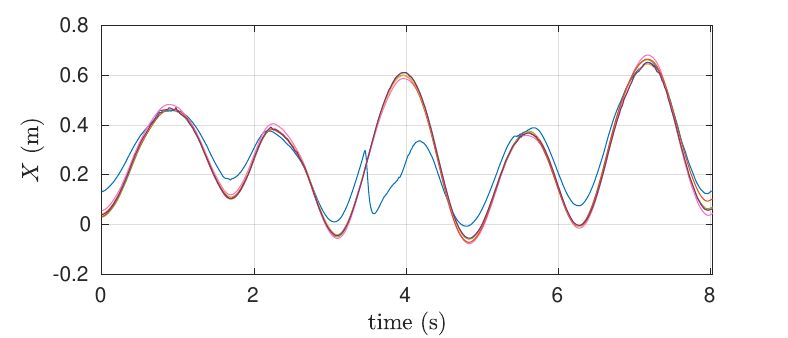}
  \end{subfigure}
  \begin{subfigure}{0.48\linewidth}
    \includegraphics[width=1.0\linewidth]{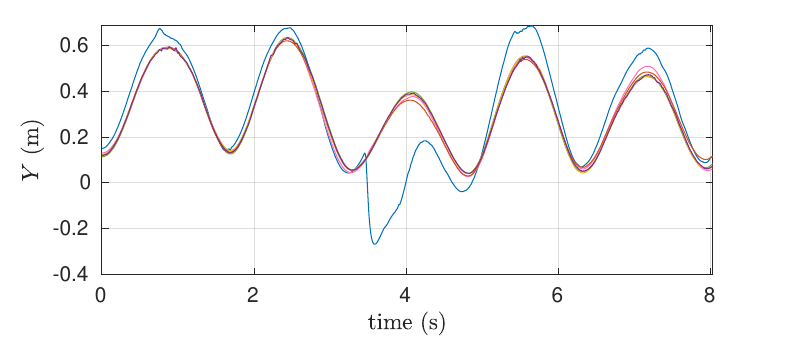}
  \end{subfigure}
  \begin{subfigure}{0.48\linewidth}
    \includegraphics[width=1.0\linewidth]{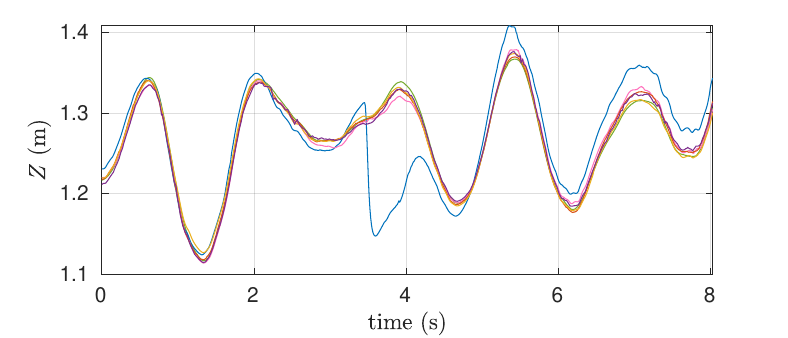}
  \end{subfigure}
  \begin{subfigure}{0.48\linewidth}
    \includegraphics[width=1.0\linewidth]{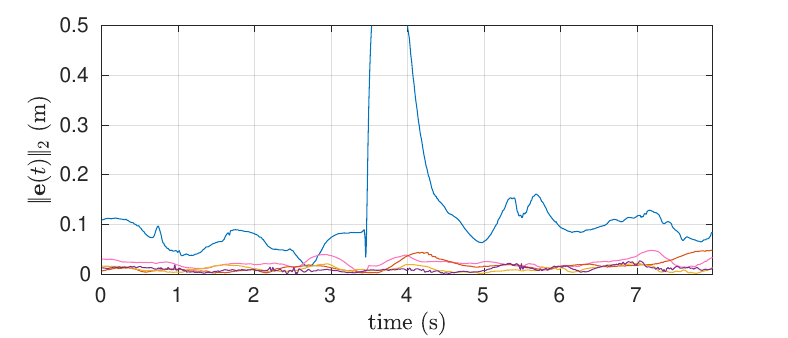}
  \end{subfigure}
  \begin{subfigure}{0.5\linewidth}
    \includegraphics[width=1.0\linewidth]{figures_supp/legend.pdf}
  \end{subfigure}
  \caption{Sequence 7 estimated trajectories versus Vicon ground truth for all available estimation methods.  All trajectories are aligned to the vicon by a rigid body transformation.  (top-left) X axis trajectory, (top-right) Y axis trajectory, (bottom-left) Z axis trajectory, (bottom-right) $l_2$ error between estimated trajectories and Vicon ground truth.}
  \label{fig:record_000006_trajectory}
\end{figure*}

\begin{figure*}
  \centering
  \begin{subfigure}{0.48\linewidth}
    \includegraphics[width=1.0\linewidth]{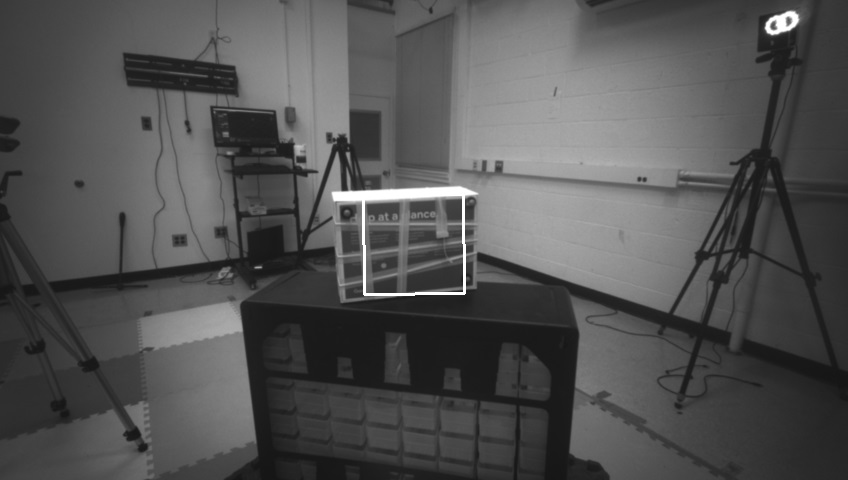}
  \end{subfigure}
  \begin{subfigure}{0.48\linewidth}
    \includegraphics[width=1.0\linewidth]{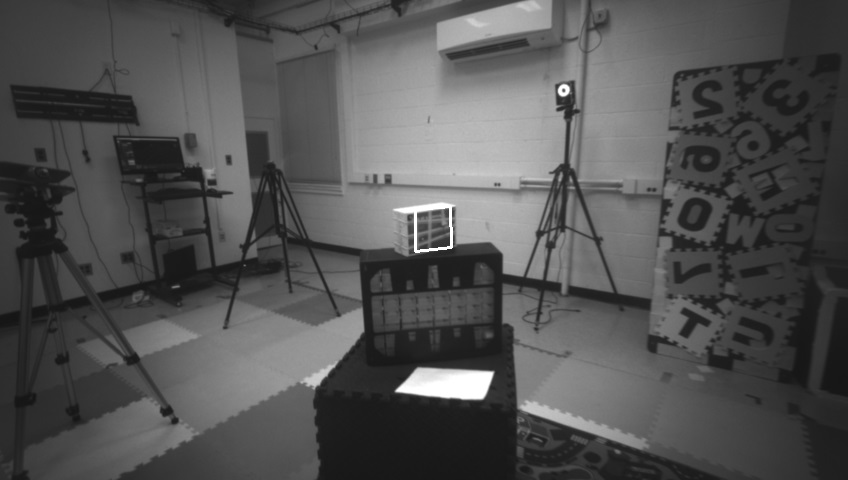}
  \end{subfigure}
  \begin{subfigure}{0.48\linewidth}
    \includegraphics[width=1.0\linewidth]{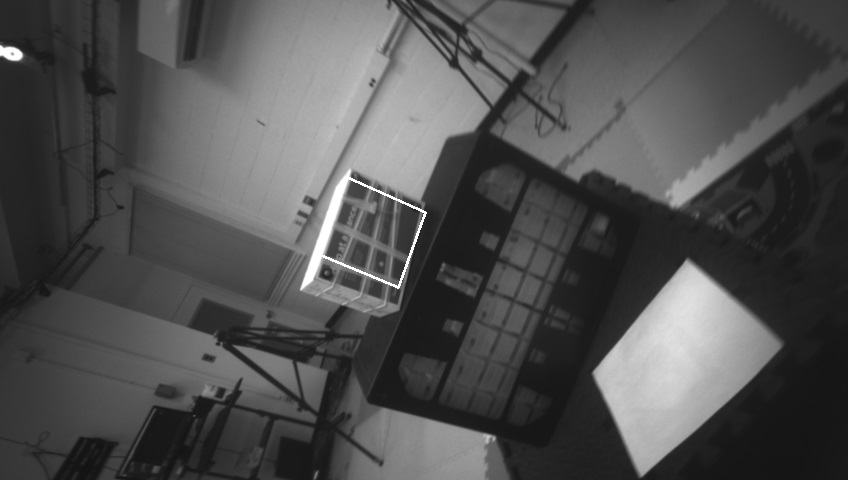}
  \end{subfigure}
  \begin{subfigure}{0.48\linewidth}
    \includegraphics[width=1.0\linewidth]{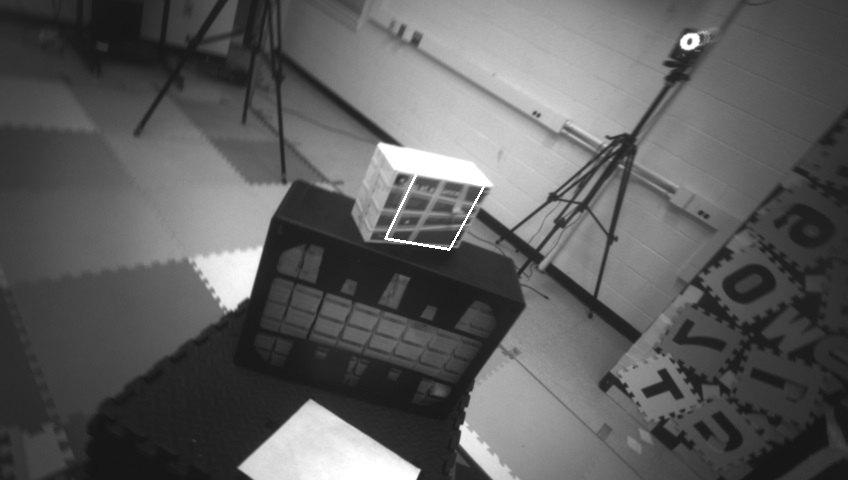}
  \end{subfigure}
  \caption{Four representative frames from Sequence 8. The tracked patch is indicated by the white box. This sequence does not contain an AprilTag.} 
\label{fig:record_000007_frames}
\end{figure*}

\begin{figure*}
  \centering
  \begin{subfigure}{0.48\linewidth}
    \includegraphics[width=1.0\linewidth]{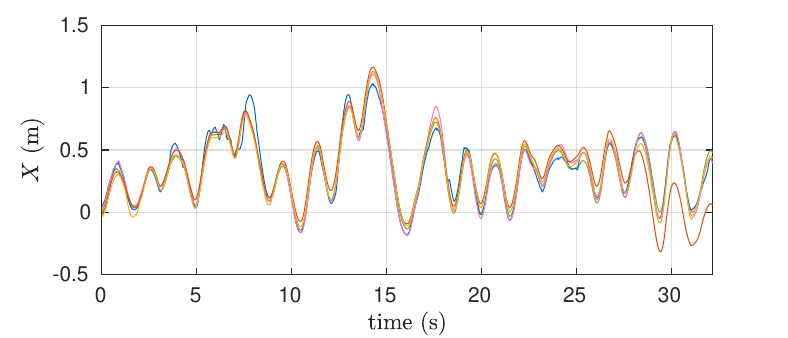}
  \end{subfigure}
  \begin{subfigure}{0.48\linewidth}
    \includegraphics[width=1.0\linewidth]{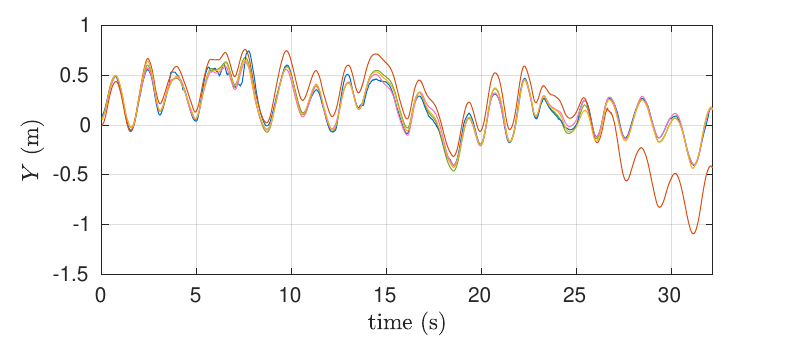}
  \end{subfigure}
  \begin{subfigure}{0.48\linewidth}
    \includegraphics[width=1.0\linewidth]{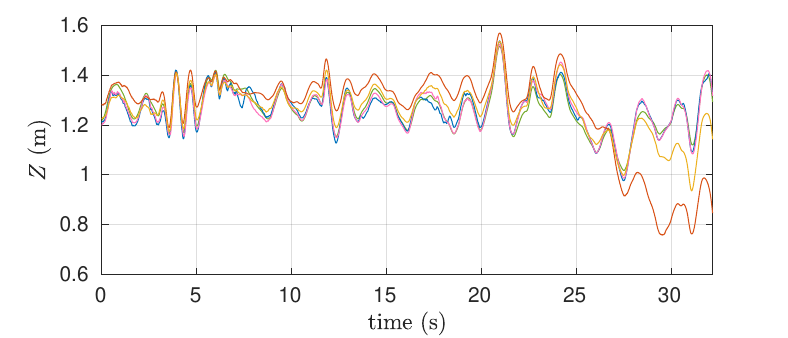}
  \end{subfigure}
  \begin{subfigure}{0.48\linewidth}
    \includegraphics[width=1.0\linewidth]{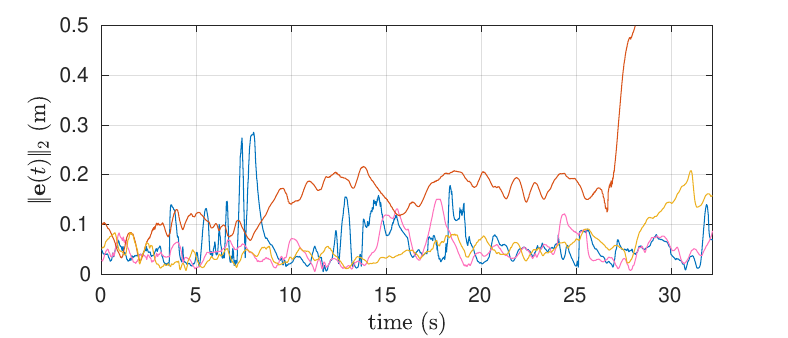}
  \end{subfigure}
  \begin{subfigure}{0.5\linewidth}
    \includegraphics[width=1.0\linewidth]{figures_supp/legend.pdf}
  \end{subfigure}
  \caption{Sequence 8 estimated trajectories versus Vicon ground truth for all available estimation methods. All trajectories are aligned to the vicon by a rigid body transformation.  (top-left) X axis trajectory, (top-right) Y axis trajectory, (bottom-left) Z axis trajectory, (bottom-right) $l_2$ error between estimated trajectories and Vicon ground truth.}
  \label{fig:record_000007_trajectory}
\end{figure*}

\begin{figure*}
  \centering
  \begin{subfigure}{0.48\linewidth}
    \includegraphics[width=1.0\linewidth]{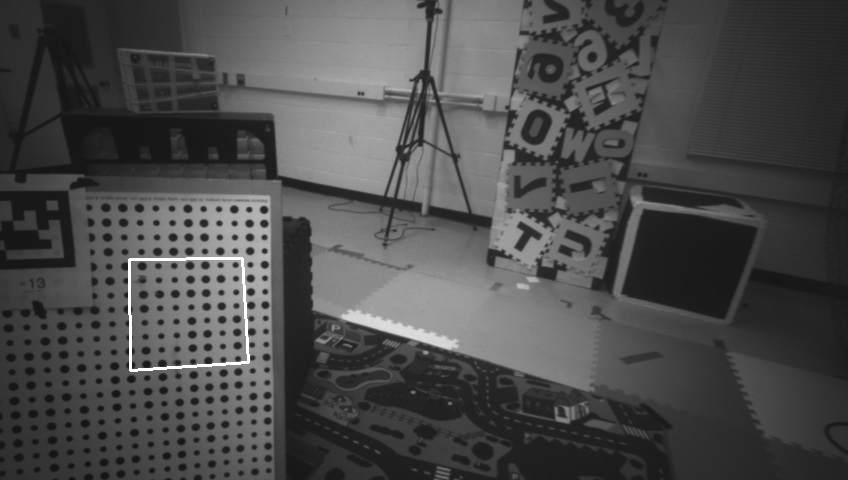}
  \end{subfigure}
  \begin{subfigure}{0.48\linewidth}
    \includegraphics[width=1.0\linewidth]{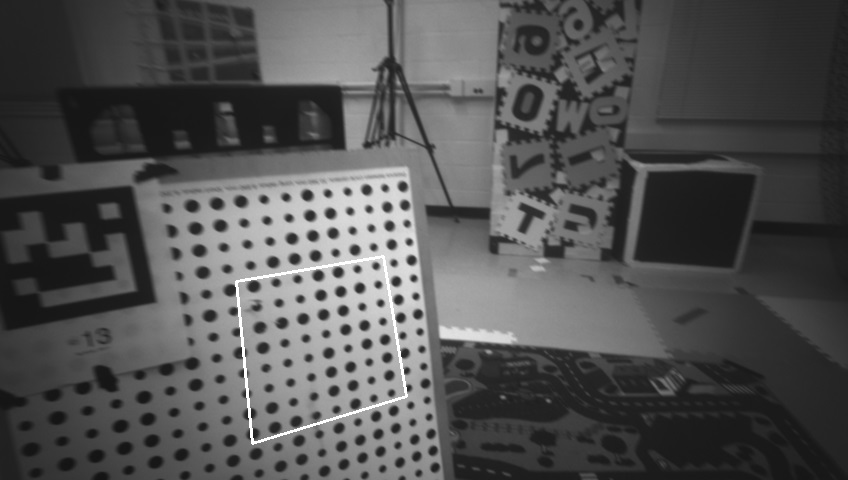}
  \end{subfigure}
  \begin{subfigure}{0.48\linewidth}
    \includegraphics[width=1.0\linewidth]{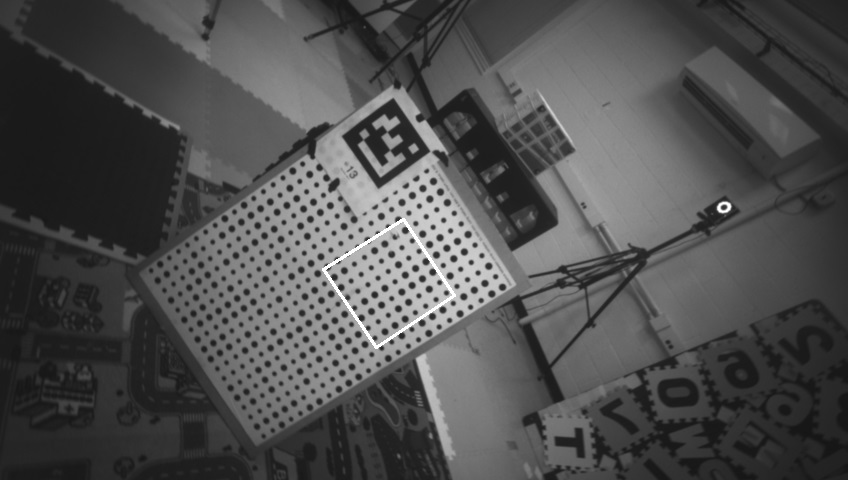}
  \end{subfigure}
  \begin{subfigure}{0.48\linewidth}
    \includegraphics[width=1.0\linewidth]{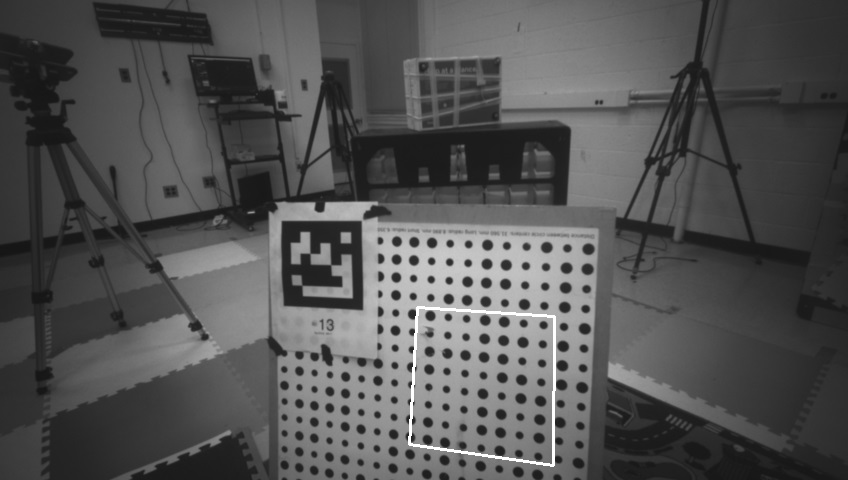}
  \end{subfigure}
  \caption{Four representative frames from Sequence 9. The tracked patch is indicated by the white box.} 
\label{fig:record_000008_frames}
\end{figure*}

\begin{figure*}
  \centering
  \begin{subfigure}{0.48\linewidth}
    \includegraphics[width=1.0\linewidth]{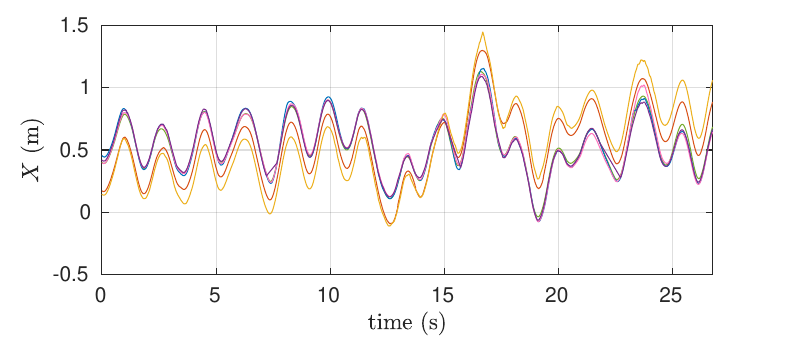}
  \end{subfigure}
  \begin{subfigure}{0.48\linewidth}
    \includegraphics[width=1.0\linewidth]{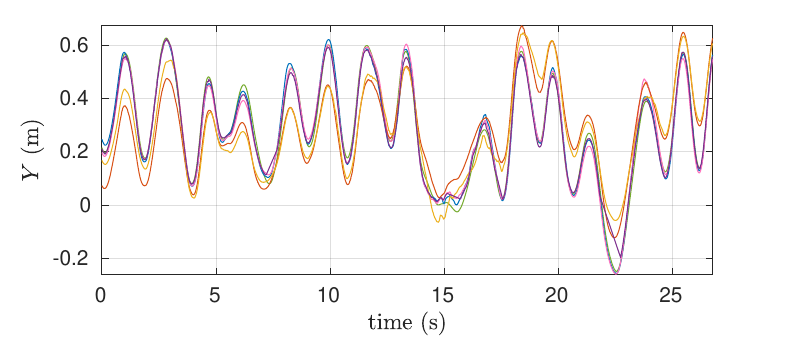}
  \end{subfigure}
  \begin{subfigure}{0.48\linewidth}
    \includegraphics[width=1.0\linewidth]{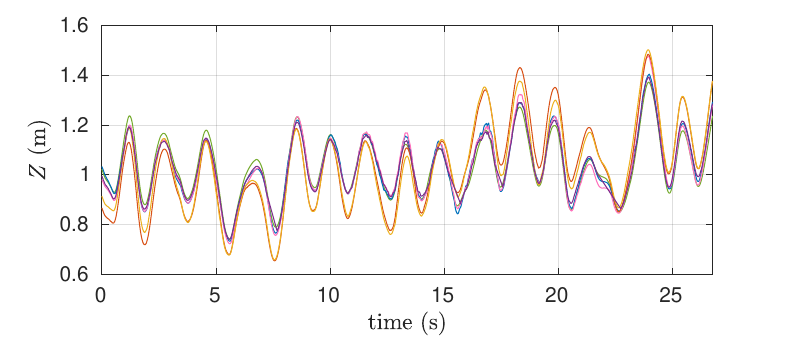}
  \end{subfigure}
  \begin{subfigure}{0.48\linewidth}
    \includegraphics[width=1.0\linewidth]{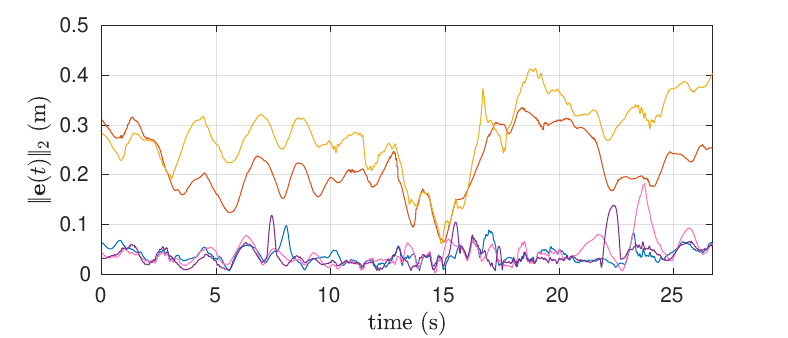}
  \end{subfigure}
  \begin{subfigure}{0.5\linewidth}
    \includegraphics[width=1.0\linewidth]{figures_supp/legend.pdf}
  \end{subfigure}
  \caption{Sequence 9 estimated trajectories versus Vicon ground truth for all available estimation methods. All trajectories are aligned to the vicon by a rigid body transformation.  (top-left) X axis trajectory, (top-right) Y axis trajectory, (bottom-left) Z axis trajectory, (bottom-right) $l_2$ error between estimated trajectories and Vicon ground truth.}
  \label{fig:record_000008_trajectory}
\end{figure*}

\begin{figure*}
  \centering
  \begin{subfigure}{0.48\linewidth}
    \includegraphics[width=1.0\linewidth]{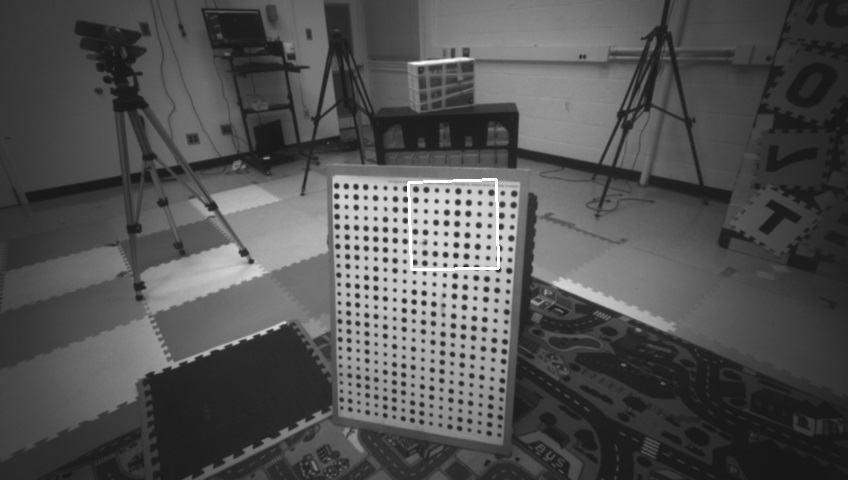}
  \end{subfigure}
  \begin{subfigure}{0.48\linewidth}
    \includegraphics[width=1.0\linewidth]{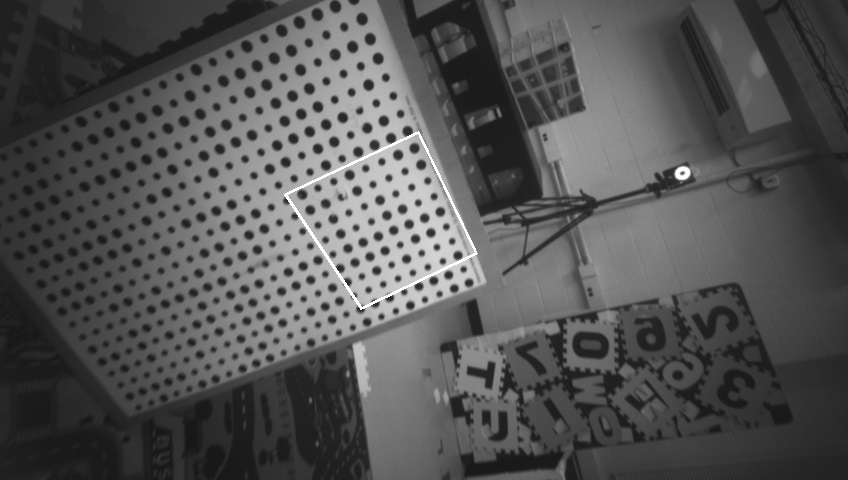}
  \end{subfigure}
  \begin{subfigure}{0.48\linewidth}
    \includegraphics[width=1.0\linewidth]{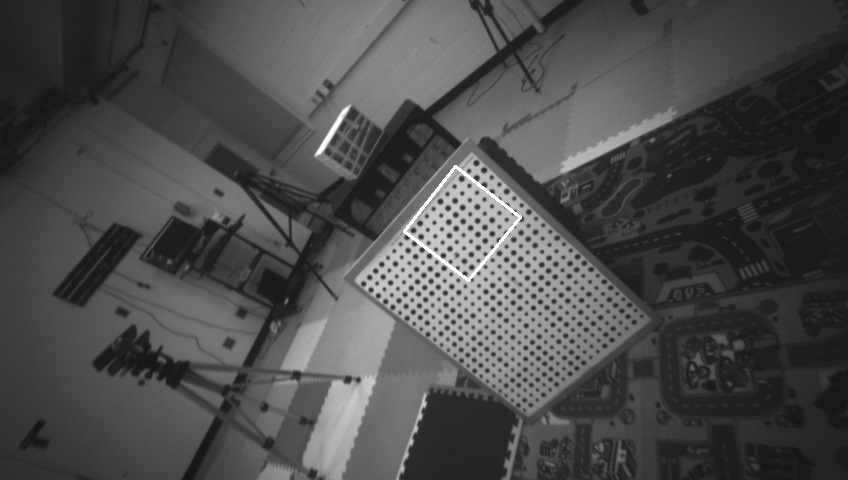}
  \end{subfigure}
  \begin{subfigure}{0.48\linewidth}
    \includegraphics[width=1.0\linewidth]{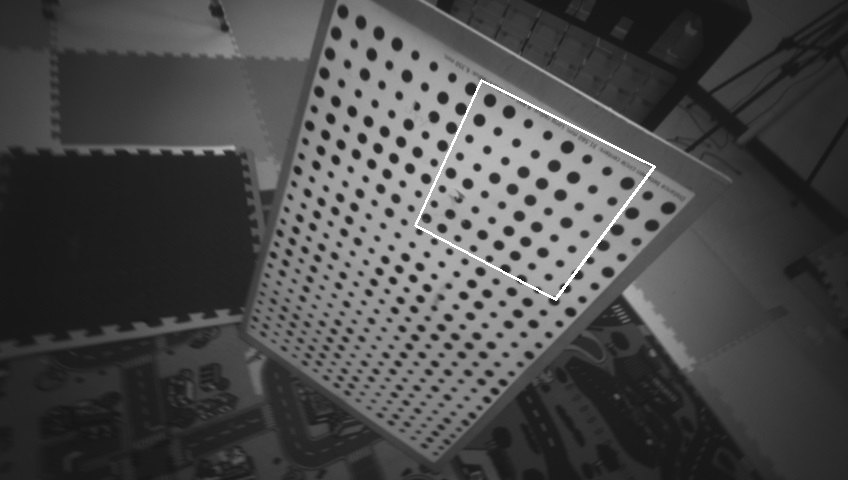}
  \end{subfigure}
  \caption{Four representative frames from Sequence 10. The tracked patch is indicated by the white box. This sequence does not contain an AprilTag.} 
\label{fig:record_000009_frames}
\end{figure*}

\begin{figure*}
  \centering
  \begin{subfigure}{0.48\linewidth}
    \includegraphics[width=1.0\linewidth]{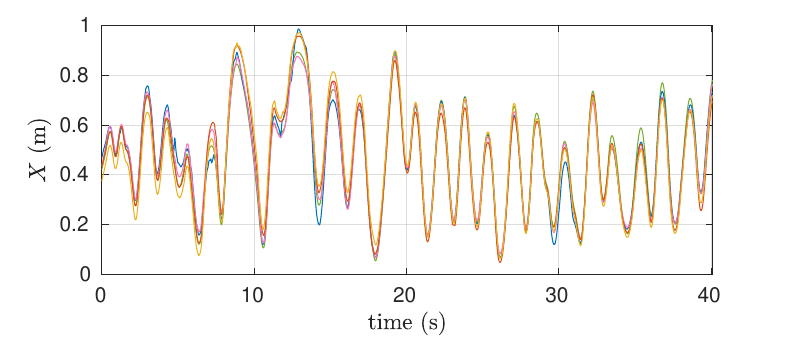}
  \end{subfigure}
  \begin{subfigure}{0.48\linewidth}
    \includegraphics[width=1.0\linewidth]{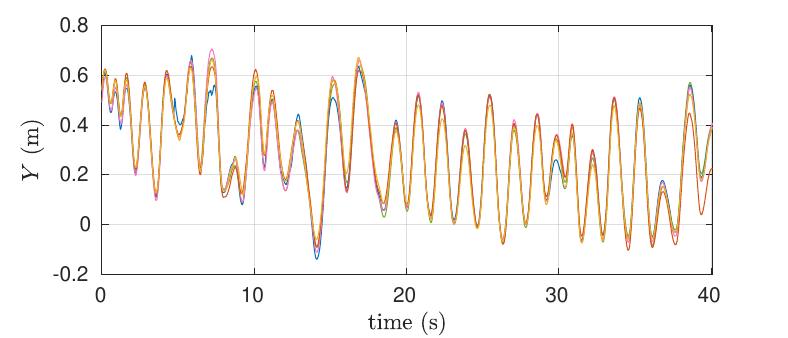}
  \end{subfigure}
  \begin{subfigure}{0.48\linewidth}
    \includegraphics[width=1.0\linewidth]{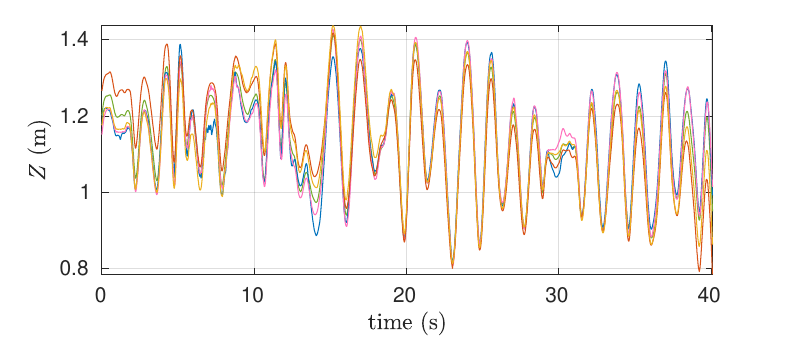}
  \end{subfigure}
  \begin{subfigure}{0.48\linewidth}
    \includegraphics[width=1.0\linewidth]{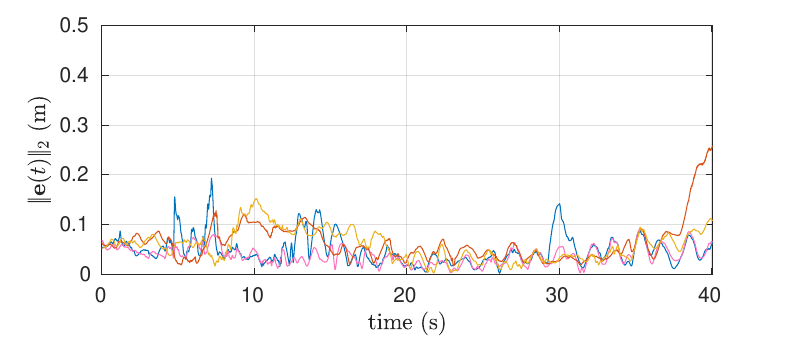}
  \end{subfigure}
  \begin{subfigure}{0.5\linewidth}
    \includegraphics[width=1.0\linewidth]{figures_supp/legend.pdf}
  \end{subfigure}
  \caption{Sequence 10 estimated trajectories versus Vicon ground truth for all available estimation methods.  All trajectories are aligned to the vicon by a rigid body transformation.  (top-left) X axis trajectory, (top-right) Y axis trajectory, (bottom-left) Z axis trajectory, (bottom-right) $l_2$ error between estimated trajectories and Vicon ground truth.}
  \label{fig:record_000009_trajectory}
\end{figure*}
\fi


